\documentclass[11pt]{article}

\usepackage[margin=1in]{geometry}
\usepackage{microtype}
\usepackage{parskip}

\usepackage{hyperref}
\hypersetup{
    colorlinks,
    linkcolor={blue!90!black},
    citecolor={magenta},
    urlcolor={blue!80!black}
}

\usepackage{gensymb} 
\usepackage{float}
\usepackage{xcolor}

\usepackage{siunitx}
\usepackage{makecell}
\usepackage{array}
\usepackage{multirow}
\usepackage{longtable}

\usepackage{microtype}
\usepackage{ragged2e}
\usepackage{wrapfig}

\usepackage{newtxtext,newtxmath} 

\usepackage{graphicx}
\usepackage{booktabs}
\usepackage{multirow}
\usepackage{array}
\usepackage{tabularx}

\usepackage{amsmath}

\usepackage{enumitem}

\usepackage{natbib}

\usepackage[font=small,labelfont=bf]{caption}
\usepackage{authblk}

\usepackage[capitalize,nameinlink,noabbrev]{cleveref}

\title{\bfseries \LARGE HydroDiffusion: Diffusion-Based Probabilistic Streamflow Forecasting with a State Space Backbone}

\author[1,2]{Yihan Wang}
\author[3]{Annan Yu}
\author[2]{Lujun Zhang}
\author[1,4]{Charuleka Varadharajan}
\author[1,4]{N. Benjamin Erichson}

\affil[1]{Lawrence Berkeley National Laboratory, Berkeley, CA, United States}
\affil[2]{University of Oklahoma, Norman, OK, United States}
\affil[3]{Cornell University, Ithaca, NY, United States}
\affil[4]{International Computer Science Institute, Berkeley, CA, United States}

\date{} 

\begin{document}
\maketitle


\begin{center}
\textbf{Abstract}
\end{center}

\begin{quotation}
\noindent
Recent advances have introduced diffusion models for probabilistic streamflow forecasting, demonstrating strong early flood-warning skill. However, current implementations rely on recurrent Long Short-Term Memory (LSTM) backbones and single-step training objectives, which limit their ability to capture long-range dependencies and produce coherent forecast trajectories across lead times. To address these limitations, we developed HydroDiffusion, a diffusion-based probabilistic forecasting framework with a decoder-only state space model backbone. The proposed framework jointly denoises full multi-day trajectories in a single pass, ensuring temporal coherence and mitigating error accumulation common in autoregressive prediction.
HydroDiffusion is evaluated across 531 watersheds in the contiguous United States (CONUS) in the CAMELS dataset. We benchmark HydroDiffusion against two diffusion baselines with LSTM backbones, as well as the recently proposed Diffusion-based Runoff Model (DRUM). Results show that HydroDiffusion achieves strong nowcast accuracy when driven by observed meteorological forcings, and maintains consistent performance across the full simulation horizon. Moreover, HydroDiffusion delivers stronger deterministic and probabilistic forecast skill than DRUM in operational forecasting. These results establish HydroDiffusion as a robust generative modeling framework for medium-range streamflow forecasting, providing both a new modeling benchmark and a foundation for future research on probabilistic hydrologic prediction at continental scales. 

\end{quotation}


\section{Introduction}
\label{sec:intro}

Deep learning has been widely used in hydrology, especially for streamflow prediction. Deep neural networks offer data-driven methods that accurately and efficiently model complex hydrological responses, and have achieved strong performance across diverse spatial and temporal prediction tasks ~\citep{kratzert2019toward,kratzert2019towards,wang2025mass,wang2025deep,liu2024probing,nearing2024global,willard2025time}.
However, most deep learning models for hydrology are deterministic, producing single point estimates, without quantifying uncertainty. Deterministic predictions are insufficient for operational predictions that must account for risk under different scenarios, such as for flood forecasting and early warning systems~\citep{krzysztofowicz2001case,pechlivanidis2025enhancing}. While ensembles of deep learning models can enable probabilistic forecasts, they are computationally intensive, requiring training of models with many different configurations~\citep{willard2025machine}. 

Diffusion-based generative models ~\citep{ho2020denoising,song2020denoising} are a promising and efficient alternative approach for probabilistic forecasting. 
These models learn a generative process that produces forecasts by transforming random noise into samples of the conditional distribution of future streamflow.
This generative process is parameterized by a neural network backbone, whose architecture and design determine a diffusion model’s capacity to approximate the conditional distribution and thus influence its overall forecasting performance.
Recently, \citet{ou2025probabilistic} introduced the Diffusion-based Runoff Model (DRUM), the first diffusion model for probabilistic hydrological forecasting. 
DRUM combines a denoising diffusion process with an encoder--decoder Long Short-Term Memory (LSTM) backbone~\citep{hochreiter1997long} to produce medium-range (up to seven-day) ensemble flood forecasts across the contiguous United States (CONUS). DRUM demonstrated strong early warning skills for extreme events. Building on this foundation, \citet{yang2025diffusion} proposed a multi-time scale LSTM diffusion model for hourly prediction (denoted h-Diffusion), showing that diffusion approaches can also extend to fine temporal resolutions and data assimilation settings.

These studies highlight the promise of diffusion-based forecasting for hydrological applications. 
However, there is an opportunity to improve time series forecasts by adopting design choices from recent advances in diffusion modeling. 
For example, DRUM and h-Diffusion rely on recurrent LSTM-based backbones that model sequences in an autoregressive manner. Therefore, they predict one step at a time and roll forward iteratively, which can lead to error accumulation and loss of temporal coherence over time. 
Moreover, DRUM uses an encoder--decoder architecture, which can introduce an information bottleneck because the encoder compresses long-range hydrologic information into a fixed representation, potentially discarding details needed for accurate forecasts~\citep{bahdanau2014neural}. 
Finally, both models are trained using the discrete denoising diffusion probabilistic model (DDPM) formulation~\citep{ho2020denoising}, where the network learns to predict noise only at fixed diffusion steps rather than along a continuous diffusion trajectory. This discretization constrains temporal continuity in the learned generative dynamics as the model is tied to a predetermined noise schedule.

In this work, we present HydroDiffusion, a diffusion-based probabilistic streamflow forecasting framework that leverages several recent key innovations.
Methodologically, we move from next-step prediction to full-sequence denoising, where the model learns the joint conditional distribution of streamflow trajectories over the entire forecast horizon. 
This formulation, now widely used in time-series diffusion models, offers advantages over autoregressive forecasting by mitigating error accumulation and enforcing temporal consistency across lead times~\citep{chen2024diffusion,yuan2024diffusion,salinas2020deepar}. 
Architecturally, we replace the LSTM backbone with a state space model (SSM)~\citep{gu2022efficiently}.  SSMs have recently been shown to capture long-range hydrologic dependencies more effectively and efficiently than LSTMs, and to achieve better performance in rainfall-runoff simulation~\citep{wang2025deep}. 
Furthermore, we adopt a decoder-only configuration, which avoids the information bottleneck and directly generates the full forecast sequence conditioned on past and future meteorological forcings as well as static catchment attributes. 
In terms of the diffusion formulation, we adopt the score-based diffusion formulation~\citep{song2020denoising} with a velocity parameterization~\citep{karras2022elucidating}, which provides a continuous-time perspective on diffusion. This formulation is known to improve both the training stability and the quality of the generated trajectories.

Our experiments target medium-range streamflow forecasting with a seven-day forecast horizon, evaluated at 531 watersheds across CONUS in the Catchment Attributes and MEteorology for Large-sample Studies (CAMELS) dataset~\citep{addor2017camels}.
We benchmark HydroDiffusion, which uses an SSM backbone, against two variants of the same diffusion framework equipped with LSTM-based backbones: (i) an encoder--decoder configuration, reflecting standard sequence-to-sequence practice; and (ii) a decoder-only configuration for direct architectural comparison to HydroDiffusion. We also include deterministic versions of the SSM and LSTM backbones to isolate the benefit introduced by the diffusion framework. In addition, we compare the performance to DRUM~\citep{ou2025probabilistic}.

In this study, we first use observation-based meteorological forcings to assess the model's ability to learn realistic watershed dynamics. We then assess operational robustness through a zero-shot reforecasting experiment, in which the trained HydroDiffusion model and the best-performing diffusion-based LSTM baseline are driven by NOAA Global Ensemble Forecast System Version 12 (GEFSv12) meteorological reforecasts~\citep{zhou2022development}.
Model performance is evaluated from initialization (Day-0) through seven-day lead times using various deterministic and probabilistic metrics. 
The results show that the proposed joint denoising formulation ensures temporal coherence, maintaining stable performance across the simulation horizon during hydrologic validation. Moreover, the choice of backbone architecture impacts the forecast skill, with the SSM-based diffusion model yielding consistent accuracy gains over LSTM counterparts in operational forecasting.
Together, these findings establish HydroDiffusion as a strong new baseline for diffusion-based hydrologic forecasting.

\section{HydroDiffusion}
\label{sec:methodology}

In this section, we first discuss the problem formulation and then the diffusion formulation that we advocate in this work. Next, we introduce the decoder-only SSM that serves as the backbone for probabilistic diffusion forecasting of medium-range streamflow, as well as the LSTM-based diffusion baselines for comparison.

\subsection{Problem Formulation of Streamflow Forecasting}
\label{subsec:problem}

The objective of this work is the cross-variable forecast of the medium-range streamflow trajectories. For each basin \(i\), our aim is to predict the nowcast of Day-0 and forecast of future streamflow (the next seven days)
\(\mathbf{q}^{(i)}_{1:L_f} = (q^{(i)}_1, q^{(i)}_2, \ldots, q^{(i)}_{L_f}) \in \mathbb{R}^{L_f \times 1}\), with \(L_f = 8\), given the past dynamic meteorological forcings 
\(\mathbf{Z}^{(i)}_{1:L_p}  (\mathbf{z}^{(i)}_1, \ldots, \mathbf{z}^{(i)}_{L_p}) \in \mathbb{R}^{L_p \times d_z}\), where $L_p=365$. Here, $d_z=5$ corresponds to precipitation, maximum and minimum temperature, shortwave radiation, and vapor pressure.

In a deterministic setting, this corresponds to learning a mapping 
\[
f: \mathbf{Z}^{(i)}_{1:L_p} \longmapsto \mathbf{q}^{(i)}_{1:L_f}.
\]

\textbf{Probabilistic Formulation.}
In the probabilistic setting, rather than predicting a single trajectory, we aim to model the conditional distribution of the streamflow given the forcings:
\[
p_\theta(\mathbf{q}^{(i)}_{1:L_f} \mid \mathbf{Z}^{(i)}_{1:L_p}),
\]
so that the model can generate multiple plausible realizations 
\(\tilde{\mathbf{q}}^{(i)}_{1:L_f} \sim p_\theta(\cdot \mid \mathbf{Z}^{(i)}_{1:L_p})\)
that reflect forecast uncertainty and hydrologic variability. 
This formulation captures both the mean behavior and the spread of possible outcomes across the forecast horizon.

To improve predictive skill and physical consistency, we extend the conditioning context to include future forcings \(\mathbf{F}^{(i)}_{1:L_{ff}} \in \mathbb{R}^{L_{ff} \times d_z}\), with $L_{ff}=7$ and static basin catchment attributes 
\(\mathbf{S}^{(i)} \in \mathbb{R}^{d_s}\),
where \(d_s = 27\) is the number of static basin characteristics. 
We define the complete conditioning tuple as
$\mathbf{c}^{(i)} := \big(\mathbf{Z}^{(i)}_{1:L_p},\; \mathbf{F}^{(i)}_{1:L_{ff}},\; \mathbf{S}^{(i)}\big)$,
and aim to approximate the conditional data distribution
\begin{equation}
p_\theta\big(\mathbf{q}^{(i)}_{1:L_f} \mid \mathbf{c}^{(i)}\big),
\label{eq:cond_target}
\end{equation}
from which samples \(\tilde{\mathbf{q}}^{(i)}_{1:L_f} \sim p_\theta(\cdot \mid \mathbf{c}^{(i)})\)
can be drawn for ensemble forecasting.
For clarity, we omit both the basin index and explicit time subscripts in the following sections and simply write \(\mathbf{q}\) and \(\mathbf{c}\) to denote the target streamflow trajectory and its conditioning context.

\subsection{Diffusion Formulation for Probabilistic Streamflow Forecasting}
\label{subsec:diffusion_model}

Having defined the probabilistic forecasting problem in \Cref{subsec:problem}, our objective is to model the conditional distribution \(p_\theta(\mathbf{q}\mid\mathbf{c})\) of future streamflow trajectories given meteorological forcings and basin attributes. Directly estimating this high-dimensional density is intractable, so instead we model its score function, \(\nabla_{\mathbf{q}} \log p_\theta(\mathbf{q}\mid\mathbf{c})\), i.e., the gradient of the log-density with respect to the data.
Specifically, we adopt the continuous-time score-based stochastic differential equation (SDE) formulation~\citep{song2020denoising} with a velocity parameterization~\citep{karras2022elucidating}, which generalizes the discrete DDPM framework~\citep{ho2020denoising}. 
This formulation defines a continuous stochastic process that provides a principled connection between the forward and reverse dynamics, yielding smoother training behavior and higher sample fidelity. 
The forward process progressively injects Gaussian noise into clean streamflow trajectories, while the reverse process, parameterized by our SSM (see \Cref{subsec:diffusion_ssm}), learns to denoise these trajectories conditioned on hydrometeorological inputs, as illustrated in \Cref{fig:diffusion_backbones}(a). 
The resulting model captures both the mean evolution and the uncertainty structure of multi-day streamflow sequences, enabling the generation of coherent and physically consistent ensemble forecasts.

\textbf{Forward process.} Diffusion introduces an auxiliary time variable \(\tau\in[0,1]\) that controls the gradual perturbation of a clean trajectory \(\mathbf{x}_0 = \mathbf{q}\) with Gaussian noise. 
Here, \(\mathbf{x}_0\) denotes the clean (noise-free) streamflow trajectory and serves as the initial condition of the diffusion process, allowing us to represent the evolving noisy samples \(\mathbf{x}_\tau\) in the same data space as \(\mathbf{q}\).
The continuous forward process is defined by the SDE
\begin{equation}
d\mathbf{x}_\tau = f(\tau)\,\mathbf{x}_\tau\,d\tau + g(\tau)\,d\mathbf{w}_\tau,
\label{eq:forward_sde}
\end{equation}
where \(\mathbf{w}_\tau\) is a standard Wiener process. The marginal at any \(\tau\) is a Gaussian of the form
\[
\mathbf{x}_\tau \;=\; \alpha_\tau\,\mathbf{x}_0 \;+\; \sigma_\tau\,\boldsymbol{\epsilon},
\]
with $\boldsymbol{\epsilon}\sim\mathcal{N}(\mathbf{0},\mathbf{I})$, and where the schedule \((\alpha_\tau,\sigma_\tau)\) satisfies \(\alpha_\tau^2+\sigma_\tau^2=1\). 
Here, the diffusion process is defined in data space, since $\boldsymbol{\epsilon}$ has the same dimension as $\mathbf{x}_0$.
Intuitively, as \(\tau\) increases, the original trajectory is progressively corrupted until it becomes nearly white noise at \(\tau=1\). 
During training, we do not simulate the full forward trajectory; instead, a single diffusion time $\tau$ is sampled uniformly from $[0,1]$, and the corresponding noisy pair $(\mathbf{x}_\tau,\tau)$ is used to train the denoising network (see Algorithm 1 in \cite{erichson2025flex}).

\textbf{Reverse (denoising) process.} To generate new trajectories, we need a mechanism that allows us to reverse this diffusion process. 
Since the forward dynamics in~\cref{eq:forward_sde} are known, we can take advantage of the fact that, under mild regularity conditions, any diffusion process with drift $f(\tau)$ and diffusion coefficient $g(\tau)$ admits a corresponding reverse-time SDE that exactly describes how to transform noise back into a clean trajectory. 
This reverse process effectively inverts the stochastic dynamics of the forward process, replacing the unknown data distribution with its evolving noisy marginal. 
Following~\citet{anderson1982reverse} and~\citet{song2020denoising}, the reverse-time SDE is given by
\begin{equation}
d\mathbf{x} \;=\; \Big[f(\tau)\,\mathbf{x} \;-\; g(\tau)^2\,\nabla_{\mathbf{x}}\log p_\tau\!\big(\mathbf{x}\mid \mathbf{c}\big)\Big]\,d\tau \;+\; g(\tau)\,d\bar{\mathbf{w}}_\tau,
\label{eq:reverse_sde}
\end{equation}
where \(\nabla_{\mathbf{x}}\log p_\tau(\mathbf{x}\mid \mathbf{c})\) is the conditional score. Conceptually, the score indicates the direction that most rapidly increases the likelihood of data and thus defines the denoising flow that reconstructs realistic streamflow trajectories. Because the true score is unknown and analytically intractable, it must be learned from observed streamflow trajectories. This is where the neural network enters: it learns an approximation of the functional representation of the score: \(\nabla_{\mathbf{x}}\log p_\tau(\mathbf{x}\mid \mathbf{c}) \approx s_\theta(\mathbf{x}_\tau,\tau,\mathbf{c})\). 

Many neural architectures have been proposed as the backbone of diffusion models. In computer vision applications, the most widely used backbone is the U-Net architecture, which offers strong performance for spatial data. However, U-Nets are not very suitable for time-series applications. The transformer backbone is a strong alternative, but it tends to require a huge amount of data for training. Recurrent networks such as LSTMs tend to be efficient for temporal modeling, yet they often have limited expressivity and struggle to capture long-range dependencies. 
To address these limitations, we propose a novel backbone based on an SSM, see \cref{subsec:diffusion_ssm},
which combines high expressivity with efficient sequence modeling capabilities, 
making it particularly well-suited for hydrologic time series.

\textbf{Velocity parameterization and training objective.}
Following~\cite{salimans2022progressive} and~\citet{karras2022elucidating}, we adopt the velocity parameterization, which is mathematically equivalent to the score up to a time-dependent scaling factor but yields improved numerical stability. For samples \((\mathbf{x}_\tau,\tau)\) from~\cref{eq:forward_sde}, the target velocity is
\begin{equation}
\mathbf{v}^\star \;=\; \alpha_\tau\,\boldsymbol{\epsilon} \;-\; \sigma_\tau\,\mathbf{x}_\tau.
\label{eq:true_velocity}
\end{equation}
We train a model \(\mathbf{v}_\theta(\mathbf{x}_\tau,\tau,\mathbf{c})\) to predict this target by minimizing the mean-squared error
\begin{equation}
\mathcal{L}(\theta)
\;=\;
\mathbb{E}_{\tau,\mathbf{x}_0,\boldsymbol{\epsilon}}
\!\left[
\big\|
\mathbf{v}_\theta(\mathbf{x}_\tau,\tau,\mathbf{c})
\;-\;
(\alpha_\tau\,\boldsymbol{\epsilon}-\sigma_\tau\,\mathbf{x}_\tau)
\big\|_2^2
\right],
\label{eq:vel_loss}
\end{equation}
with \(\tau\) sampled uniformly over \([0,1]\). As shown by~\cite{erichson2025flex}, this parameterization reduces the variance of the learned velocity field, improving both training stability and sample quality. Each model output is a length-\(L_f\) vector, representing the full trajectory.

\textbf{Sampling procedure.}
Once trained, conditional samples are generated by integrating the reverse dynamics from a noise realization \(\mathbf{x}_1\!\sim\!\mathcal{N}(\mathbf{0},\mathbf{I})\) to \(\mathbf{x}_0\) using the deterministic denoising diffusion implicit model (DDIM) sampler~\citep{song2020denoising}.  
The DDIM formulation can be interpreted as integrating the probability flow ordinary differential equation (ODE) associated with the reverse-time SDE in~\cref{eq:reverse_sde}, describing a deterministic trajectory that shares the same marginal probability evolution as the stochastic reverse process.
Using this approach, sampling proceeds via the following update:
\begin{equation}
\mathbf{x}_{\tau_{t-1}} = 
\alpha_{\tau_{t-1}}\,\mathbf{x}_{0} 
+ \sigma_{\tau_{t-1}}\,\mathbf{v}_{\tau_t}.
\label{eq:ddim_update_core}
\end{equation}
At each discrete index \(t\), corresponding to a diffusion time 
\(\tau_t \in [0,1]\),
the model predicts the velocity field
\[
\mathbf{v}_{\tau_t} = 
\mathbf{v}_\theta(\mathbf{x}_{\tau_t}, \tau_t, \mathbf{c}),
\]
and computes the clean-sample estimate under the current variance schedule 
\((\alpha_{\tau_t}, \sigma_{\tau_t})\):
\begin{equation}
\mathbf{x}_{0} = 
\alpha_{\tau_t}\,\mathbf{x}_{\tau_t} 
- \sigma_{\tau_t}\,\mathbf{v}_{\tau_t}.
\label{eq:x0_from_v}
\end{equation}
The update in~\cref{eq:ddim_update_core} is applied iteratively, starting from the final diffusion step corresponding to \(\tau_T = 1\) and proceeding backward through decreasing diffusion times until the initial step \(\tau_1 \approx 0\).
The discrete indices \(t \in \{1, \dots, T\}\) correspond to a uniform discretization
of the continuous diffusion time \(\tau \in [0,1]\), with \(\tau_t = t/T\).
In practice, the number of steps \(T\) can be adjusted to balance sample quality and computational cost.
Because the velocity parameterization provides a smooth and stable reverse trajectory,
high-quality samples can be obtained with as few as ten discrete steps in our setup.

\subsection{A State Space Model Backbone}
\label{subsec:diffusion_ssm}

Having specified the diffusion formulation, we now introduce an SSM that parameterizes the conditional velocity field, which governs the reverse-time dynamics of the diffusion process. 
At each reverse step, the model receives the current noisy forecast sequence $\mathbf{x}_\tau$, the diffusion time $\tau$, and the conditioning inputs $\mathbf{c}$, and predicts the corresponding velocity $\mathbf{v}_\theta(\mathbf{x}_\tau, \tau, \mathbf{c})$ that directs $\mathbf{x}_\tau$ toward the clean streamflow trajectory in physical time. 
This formulation allows the model to learn the temporal evolution of denoising in a physically consistent manner, rather than predicting individual streamflow values directly.

Learning the conditional velocity field in the diffusion framework requires a backbone that is both expressive and scalable, capable of representing complex temporal dependencies across long hydrologic records. 
Recurrent networks such as LSTMs, though widely used in hydrologic modeling, process sequences sequentially and often suffer from limited temporal context, vanishing gradients, and poor parallel scalability. 
In contrast, SSMs represent temporal dynamics using continuous-time linear systems. This formulation allows the SSMs to be discretized at an arbitrarily fine resolution, which effectively mitigates gradient vanishing issues and enables robust modeling of long-range dependencies. Furthermore, SSMs fundamentally enable efficient sequence modeling by computing the full sequence in parallel via a convolutional representation. 

\textbf{State Space Formulation.}
In continuous time, an SSM can be described as a linear time-invariant system~\citep{gu2022efficiently}:
\begin{align}
\mathbf{h}'(t) &= \mathbf{A}\mathbf{h}(t) + \mathbf{B}\mathbf{u}(t), \label{eq:contLTI_state} \\
\mathbf{z}(t) &= \mathbf{C}\mathbf{h}(t) + \mathbf{D}\mathbf{u}(t), \label{eq:contLTI_output}
\end{align}
where $\mathbf{h}(t)\!\in\!\mathbb{C}^{n}$ denotes the latent state, $\mathbf{u}(t)\!\in\!\mathbb{C}^{m}$ the input, and $\mathbf{z}(t)\!\in\!\mathbb{C}^{p}$ the output. 
The matrices $\mathbf{A},\mathbf{B},\mathbf{C},\mathbf{D}$ govern the system dynamics and are learned during training.

For practical implementation, ~\cref{eq:contLTI_state} and ~\cref{eq:contLTI_output} need to be discretized and parameterized with a learnable sampling interval $\Delta t$, which allows the model to adaptively represent processes occurring at different characteristic timescales:
\begin{equation}
\mathbf{h}_k = \overline{\mathbf{A}}\mathbf{h}_{k-1} + \overline{\mathbf{B}}\mathbf{u}_k,\qquad
\mathbf{z}_k = \overline{\mathbf{C}}\mathbf{h}_{k} + \overline{\mathbf{D}}\mathbf{u}_k,
\end{equation}
where $\overline{\mathbf{A}},\overline{\mathbf{B}},\overline{\mathbf{C}},\overline{\mathbf{D}}$ are the discretized parameters induced by $(\mathbf{A},\mathbf{B},\mathbf{C},\mathbf{D},\Delta t)$. 
Here, $k$ indexes the discrete time steps of the SSM’s internal dynamics as determined by $\Delta t$. 
Unrolling this recurrence yields a 1D convolution in physical time between the input sequence $\mathbf{u}$ and a kernel $\overline{\mathbf{K}}$:
\begin{equation}
\mathbf{z} = \mathbf{u} * \overline{\mathbf{K}}, \qquad
\overline{\mathbf{K}} = \big[(\overline{\mathbf{C}}\overline{\mathbf{B}}), (\overline{\mathbf{C}}\overline{\mathbf{A}}\overline{\mathbf{B}}), \ldots, (\overline{\mathbf{C}}\overline{\mathbf{A}}^{L-1}\overline{\mathbf{B}})\big],
\end{equation}
which can be efficiently computed in the frequency domain using Fast Fourier Transforms (FFTs). This parallel formulation enables efficient processing of long sequences.

Here, we adopt the frequency-tuned diagonal SSM (S4D-FT)~\citep{yu2024tuning}. S4D-FT builds on the diagonal state-space formulation of the structured state space (S4D) model~\citep{gu2022parameterization}, which decomposes the dynamics into independent single-input/single-output channels. While S4D provides efficient and stable long-range modeling, it tends to underrepresent high-frequency variations that are essential for hydrologic forecasting, such as rapid runoff or convective storm responses.

S4D-FT addresses this limitation by introducing a learnable frequency-tuning mechanism that adjusts the spectral characteristics of each channel. 
Let $\mathbf{A} = \mathrm{diag}(\lambda_1, \ldots, \lambda_n)$ with $\lambda_j \in \mathbb{C}$. 
The base eigenvalues $\{\lambda^{\text{base}}_j\}$ are initialized in log-polar coordinates and scaled by learnable real parameters $(\alpha_r, \alpha_i)$:
\begin{equation}
\lambda_j = \alpha_r\,\mathrm{Re}(\lambda^{\text{base}}_j) + i\,\alpha_i\,\mathrm{Im}(\lambda^{\text{base}}_j).
\end{equation}
Here, $\alpha_r$ controls low-frequency memory decay, while $\alpha_i$ modulates oscillatory behavior and high-frequency response. This frequency-tuning flexibility enables the model to jointly represent both slow (e.g., groundwater and snowmelt processes) and fast (e.g., precipitation events) hydrologic dynamics within a unified architecture.

Recent work by \citet{wang2025deep} demonstrated that S4D-FT outperforms both LSTM and standard S4D variants in large-sample hydrologic modeling. Such success makes S4D-FT a natural choice as the backbone for HydroDiffusion, enabling high-fidelity denoising and ensemble generation.

\begin{figure}[!t]
\centering
\includegraphics[width=\textwidth]{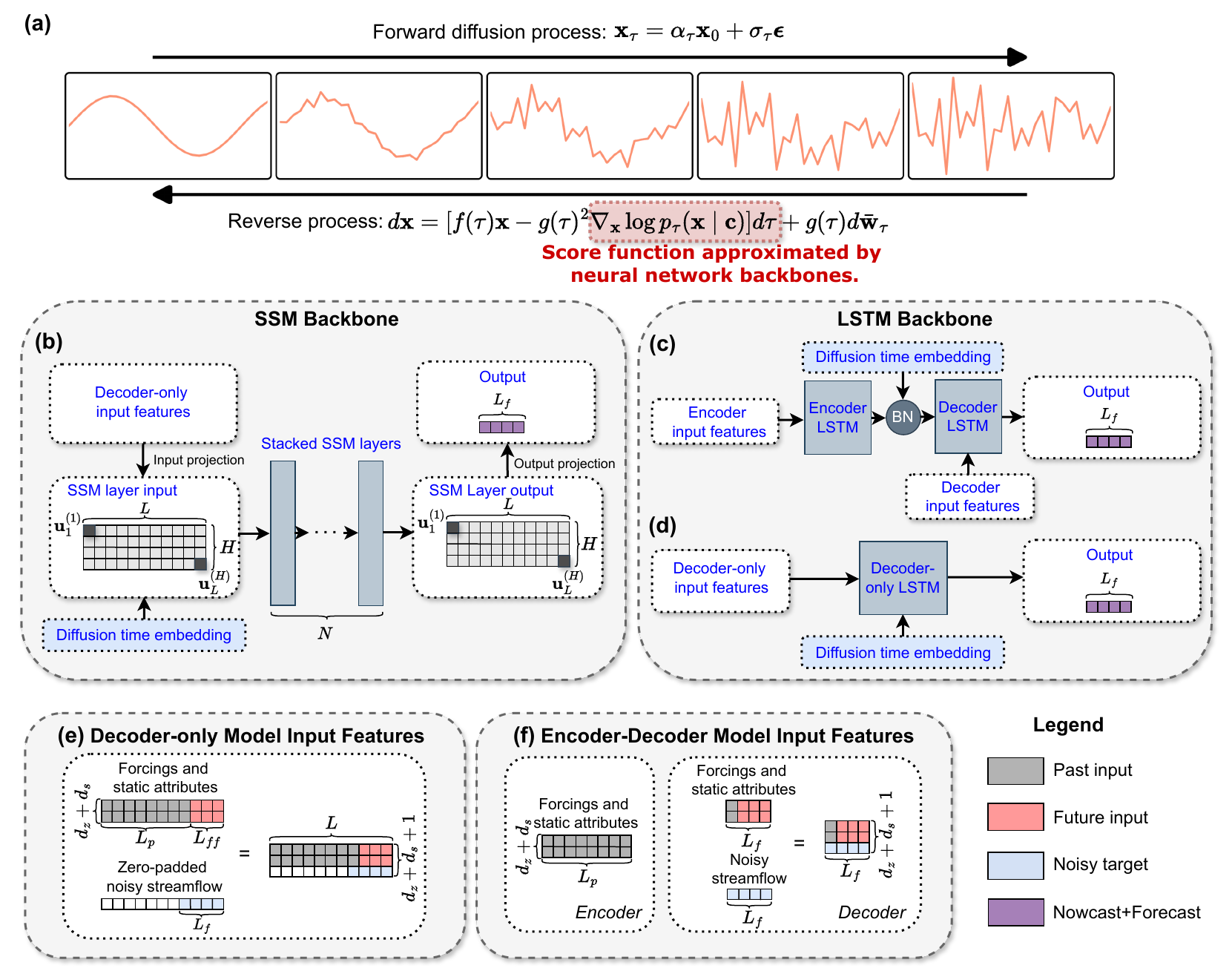}
\caption{
Schematic of the diffusion framework and neural network backbones for streamflow forecasting. 
(a) Forward diffusion and reverse denoising processes, where the score function is approximated by neural network backbones. 
(b) \textbf{HydroDiffusion} with an SSM backbone (specifically S4D-FT), which processes the full input sequence using $N$ stacked SSM layers with $H$ latent state channels, together with input/output projections and diffusion-time embedding. 
(c) \textbf{DiffusionLSTM\textsubscript{EncDec}} with an encoder--decoder structure: the encoder LSTM summarizes historical inputs, and the decoder LSTM generates future streamflow conditioned on future forcings and diffusion-time embedding. 
(d) \textbf{DiffusionLSTM\textsubscript{Dec}} with a decoder-only structure that directly maps the combined past and future inputs to streamflow forecasts. 
(e--f) Input feature configurations for decoder-only and encoder--decoder models, respectively, showing how forcings, static attributes, and noisy streamflow targets are arranged. 
Gray, red, blue, and purple grids denote past inputs, future inputs, noisy targets, and nowcast+forecast outputs, respectively. 
The number of grids is illustrative and does not reflect the actual sequence length or lead time used in this study.
}

\label{fig:diffusion_backbones}
\end{figure}

\Cref{fig:diffusion_backbones}(b) illustrates the SSM backbone. At each reverse diffusion step~$\tau$, the network receives the current noisy streamflow sequence~$\mathbf{x}_\tau$, the diffusion time~$\tau$, and the conditioning context.
The construction of the model input sequence is detailed in \Cref{fig:diffusion_backbones}(e):
past and future meteorological forcings~$\mathbf{Z}$, static basin attributes~$\mathbf{S}$, and the noisy forecast trajectory~$\mathbf{x}_\tau$ are concatenated along the feature dimension. 
The noisy trajectory is zero-padded over the past window to align with the full conditioning length~$L$.
All inputs are projected into a shared $H$-dimensional latent space. 
A Fourier-based diffusion-time embedding of~$\tau$ is injected into each SSM layer as an additive bias over the nowcast and forecast window $L_f$, enabling the model to adapt its denoising behavior to the current noise level. 
The embedded sequence is then processed by~$N$ stacked S4D-FT layers with residual connections and nonlinear activations, which mix information across time through frequency-tuned convolutions. 
Finally, an output projection maps the resulting latent representation to a length-$L$ sequence, and the final length-$L_f$ elements are used as a streamflow forecast. 
By denoising the entire~$\mathbf{x}_\tau$ sequence jointly (non-autoregressively), we learn the conditional distribution of the full trajectory, ensuring temporal coherence and preventing error accumulation across lead times.

For comparison, \Cref{fig:diffusion_backbones}(c) and (d) illustrate the two LSTM-based architectures, which we collectively refer to as DiffusionLSTM. DiffusionLSTM is used to contrast with the proposed SSM backbone in HydroDiffusion; the underlying score-based diffusion formulation remains unchanged. The encoder--decoder variant (DiffusionLSTM\textsubscript{EncDec}; panel~c) encodes historical forcings and static attributes over the lookback window $L_p$ into a latent bottleneck (hidden and cell states). 
Diffusion-time embeddings are injected into this bottleneck as additive bias, and the decoder generates the denoised streamflow sequence over $L_f$ conditioned on future meteorological inputs and static attributes. The construction of input sequences for this encoder--decoder architecture is detailed in \Cref{fig:diffusion_backbones}(f).
The decoder-only variant (DiffusionLSTM\textsubscript{Dec}; panel~d) removes the encoder and concatenates past and future forcings, static attributes, and a zero-padded noisy sequence $\mathbf{x}_\tau$ into a single decoder input. 
Here, the diffusion-time embedding is added at the sequence boundary to condition the model explicitly on $\tau$. 
Together, these LSTM-based baselines provide direct structural comparisons to the SSM backbone, isolating the influence of the backbone architecture on diffusion-based sequence modeling.

\section{Experimental Setup}
\label{sec:experimental_setup}
In this section, we first introduce the dataset. We then present the design rationale of the proposed two-phase, diffusion-based forecasting framework. Lastly, we describe the model training configurations, evaluation metrics, and evaluation procedure used to assess the effectiveness of HydroDiffusion against several baseline models.

\subsection{Dataset}
\label{sec:dataset}
The CAMELS dataset used in this study provides long-term hydrometeorological records and basin characteristics across CONUS. 
CAMELS includes multiple meteorological forcing products, including the North American Land Data Assimilation System (NLDAS; \citealt{xia2012continental}), the Maurer dataset~\citep{maurer2002long}, and Daymet~\citep{thornton1997generating}. To compare with ~\citet{ou2025probabilistic}, we use five dynamic meteorological variables from Daymet as forcings (precipitation, maximum and minimum temperature, shortwave radiation, and vapor pressure) together with 27 static watershed attributes (listed in \Cref{tab:variables} of the Appendix).
A total of 531 watersheds are selected after excluding basins larger than 2,000~km$^2$ and those with inconsistent drainage area estimates across products. 
This subset has been widely used in large-scale streamflow modeling studies~\citep{frame2023strictly,wang2025deep,ou2025probabilistic}.

For the operational forecasting experiments, we employ the NOAA GEFSv12 reforecast product, which provides global reforecasts of near-surface meteorological variables at 0.25° spatial resolution with 31 ensemble members issued every three hours. 
We select the same five dynamic meteorological variables as used from Daymet. 
For each variable, the control member is extracted, bilinearly interpolated to 0.125°, basin-averaged, and aggregated daily to generate the model inputs. 
Because vapor pressure ($e$, Pa) is not directly available in GEFSv12, it is derived from specific humidity ($q$, kg~kg$^{-1}$) and surface air pressure ($p$, Pa) using the standard relation 
$e = \frac{q\,p}{0.622 + 0.378\,q}$.

\subsection{A Two-Phase Diffusion-Based Forecasting Framework}
\label{subsec:forecast_framework}

\begin{wrapfigure}{r}{0.49\textwidth}
\centering
\vspace{-0.9cm}
\includegraphics[width=0.49\textwidth]{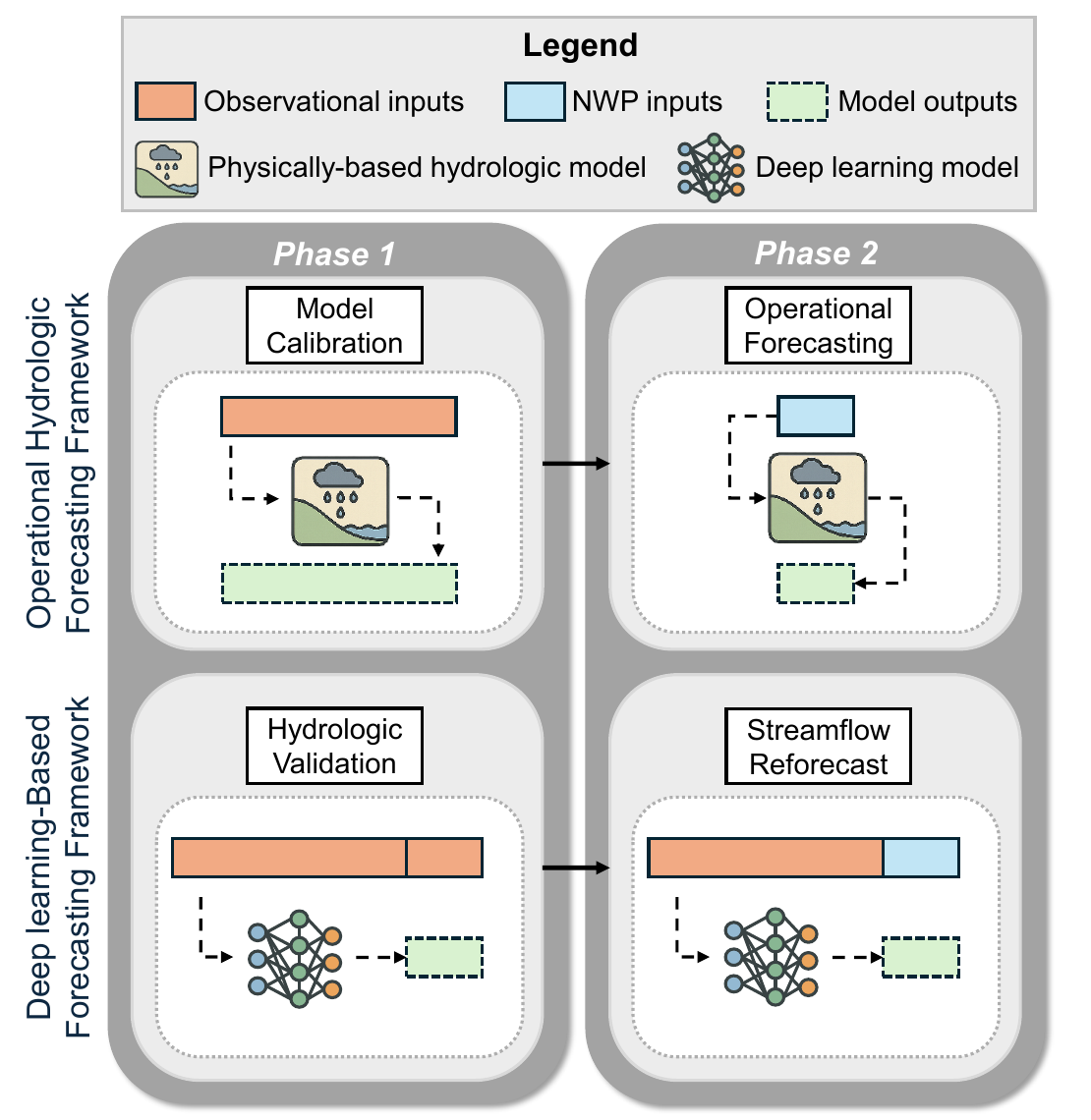}
\caption{Comparison between the operational hydrologic forecasting framework (top row) and the proposed deep learning-based forecasting framework (bottom row). }
\vspace{-0.3cm}
\label{fig:forecast_framework_scheme}
\end{wrapfigure}

The experimental design of this study follows the logic of operational hydrologic forecasting, which typically consists of two sequential phases: model calibration and operational forecasting, as illustrated in the top row of \Cref{fig:forecast_framework_scheme}. 
In the calibration phase, a physically-based hydrologic model is tuned using observed hydrometeorological data from a historical period to ensure that it can adequately reproduce watershed dynamics~\citep{moriasi2007model}. 
In the forecasting phase, the calibrated model is driven by numerical weather prediction (NWP) forcings to produce real-time streamflow forecasts.

To mirror this workflow within a deep learning framework for physical and operational plausibility, our study adopts two experimental phases: \emph{hydrologic validation} and \emph{streamflow reforecast}, shown in the bottom panel of \Cref{fig:forecast_framework_scheme}.
In the hydrologic validation phase, HydroDiffusion and two DiffusionLSTM baselines are trained using historical Daymet observations in a 365-to-8 sequence-to-sequence setup, where each model output spans the Day-0 and seven-day simulation horizon (as defined in~\cref{subsec:problem}). 
This phase evaluates whether the diffusion models can learn watershed hydrologic dynamics under the proposed joint denoising strategy.
In the streamflow reforecast phase, the trained models are driven by meteorological reforecasts from the GEFSv12 dataset. 
Inputs combine two segments: observed meteorological forcings prior to (and including) initialization and GEFSv12 reforecasts thereafter, closely following operational hydrologic forecasting practice.

During hydrologic validation, we include an LSTM-based streamflow simulation benchmark across CONUS to establish a reliable reference. Although several LSTM-based benchmarks exist for CONUS (e.g.,~\citealp{frame2023strictly,wang2025deep}), none match the forcing data and temporal splits used for the diffusion models. To ensure a fair comparison, we reproduce an LSTM benchmark under the same Daymet forcings and time periods.
The reproduced benchmark achieves accuracy consistent with previously reported studies (\cref{tab:day0_main}), confirming its suitability as a reference standard. Using this simulation benchmark, we evaluate whether the diffusion models achieve comparable performance from initialization through the seven-day horizon, which would indicate realistic hydrologic behavior and readiness for reforecast experiments. We also compare against DRUM, the first and only existing diffusion-based framework for medium-range hydrologic forecasting.

Following validation, we assess the reforecast performance of HydroDiffusion relative to its best-performing LSTM-based counterpart, using deterministic and probabilistic metrics over seven-day lead times and spatial domains. 
Together, these analyses provide a comprehensive assessment of diffusion models with different backbone architectures and highlight their respective strengths and limitations.

\subsection{Model Training Setup and Computational Cost}
\label{sec:training}

Following~\citet{ou2025probabilistic}, all diffusion models are trained over 10/1/1980--9/30/1990 and validated over 10/1/1990--9/30/1995. 
For the hydrologic validation phase, model performance is evaluated on 10/1/1995--9/30/2005. 
For the streamflow reforecast phase, the evaluation period is 10/1/2000--9/30/2005, as GEFSv12 reforecasts are available from 2000 onward.

All 32 input and target variables are normalized using the mean and standard deviation computed from the training period. 
HydroDiffusion and DiffusionLSTM are trained for 60~epochs using the Lion optimizer~\citep{chen2023symbolic} with a base learning rate of $3\times10^{-5}$. 
The learning rate follows a linear warm-up during the first epoch and a linear decay schedule thereafter. 
HydroDiffusion adopts the final training epoch (epoch~60) for evaluation, whereas DiffusionLSTM uses the checkpoint with the lowest validation loss, as using the final epoch leads to degraded performance. 
All model hyperparameters are manually tuned based on those used in ~\citet{wang2025deep}, with final values reported in~\Cref{tab:lstm_hparams} and~\Cref{tab:hydrodiffusion_hparams} of the Appendix. 
During inference, each diffusion model generates an ensemble of 50~members for each watershed and lead time, following~\citet{ou2025probabilistic}. In this study, we use ten DDIM sampling steps for generating forecasts.

To isolate the contribution of the diffusion-based generative mechanism, we also train a deterministic version for all diffusion models (HydroDiffusion and DiffusionLSTM variants) in the same 365-to-8 (including Day-0) sequence-to-sequence setup. 
The hyperparameters for all deterministic models follow~\citet{wang2025deep}.
Full configuration details are provided in~\Cref{tab:lstm_hparams} and~\Cref{tab:ssm_hparams} of the Appendix.

In terms of computational cost, HydroDiffusion requires approximately 20~minutes per training epoch on a single NVIDIA~L40S~GPU, while DiffusionLSTM\textsubscript{EncDec} and DiffusionLSTM\textsubscript{Dec} require about 18~and~15~minutes per epoch, respectively. During inference, the decoder-only architectures are substantially more demanding: HydroDiffusion requires roughly~50~hours to generate forecasts for all~531~basins, and DiffusionLSTM\textsubscript{Dec} about~30~hours, whereas DiffusionLSTM\textsubscript{EncDec} completes the same task in roughly~4~hours. 
This difference arises because, during inference, the encoder--decoder architecture executes only the lightweight decoder, while decoder-only models must process the entire historical input sequence at each diffusion step, resulting in significantly higher computational cost.

\subsection{Evaluation}
\label{sec:evaluation}

\subsubsection{Evaluation Metrics}

Model performance is assessed using both deterministic and probabilistic metrics to comprehensively evaluate accuracy, bias, and uncertainty representation in streamflow forecasting. 
For deterministic evaluation, we employ the Nash–Sutcliffe Efficiency (NSE; \citealp{nash1970river}), Kling–Gupta Efficiency (KGE; \citealp{gupta2009decomposition}), correlation coefficient (COR), and the percent bias in the flow-duration-curve high-flow (FHV) and low-flow (FLV) segments~\citep{yilmaz2008process}. 
The corresponding formulations are given in \Cref{eq:nse}–\Cref{eq:flv}:
\begin{equation}
\mathrm{NSE} = 1 - \frac{\sum_t (Q_t^{\mathrm{obs}} - Q_t^{\mathrm{sim}})^2}{\sum_t (Q_t^{\mathrm{obs}} - \overline{Q^{\mathrm{obs}}})^2},
\label{eq:nse}
\end{equation}
\begin{equation}
\mathrm{COR} = 
\frac{\mathrm{cov}(Q^{\mathrm{sim}}, Q^{\mathrm{obs}})}{\sigma(Q^{\mathrm{sim}})\,\sigma(Q^{\mathrm{obs}})},
\label{eq:cor}
\end{equation}
\begin{equation}
\mathrm{KGE} = 1 - \sqrt{
\left( 
COR - 1 
\right)^{2}
+ 
\left( 
\frac{\sigma_{sim}}{\sigma_{obs}} - 1 
\right)^{2}
+ 
\left( 
\frac{\overline{Q^{\mathrm{sim}}}}{\overline{Q^{\mathrm{obs}}}} - 1 
\right)^{2}
},
\label{eq:kge}
\end{equation}
\begin{equation}
\mathrm{FHV (\%)} = 100 \times \frac{\sum_{i=1}^{n_h} (Q_{(i)}^{\mathrm{sim}} - Q_{(i)}^{\mathrm{obs}})}{\sum_{i=1}^{n_h} Q_{(i)}^{\mathrm{obs}}},
\label{eq:fhv}
\end{equation}
\begin{equation}
\mathrm{FLV(\%)} = 100 \times \frac{\sum_{i=n-n_l+1}^{n} (Q_{(i)}^{\mathrm{sim}} - Q_{(i)}^{\mathrm{obs}})}{\sum_{i=n-n_l+1}^{n} Q_{(i)}^{\mathrm{obs}}}.
\label{eq:flv}
\end{equation}

Here, $Q_t^{\mathrm{obs}}$ and $Q_t^{\mathrm{sim}}$ denote observed and simulated streamflow at time $t$, $\overline{Q}$ and $\sigma(Q)$ represent temporal mean and standard deviation, and $\mathrm{cov}(\cdot)$ is covariance. 
The ranked flows $Q_{(i)}$ are ordered from highest to lowest, $n$ is the total number of samples, and $n_h$ and $n_l$ are the numbers of samples in the high-flow (top~0.1\%) and low-flow (bottom~30\%) segments. 
Optimal values for NSE, COR, and KGE are~1, while those for FHV and FLV are~0.

For probabilistic evaluation, we use the Continuous Ranked Probability Score (CRPS; \citealp{gneiting2007strictly}), complemented by reliability, sharpness, and precision–recall (PR) analyses that evaluate the model’s ability to predict the top 10\% high-flow exceedance events. 
CRPS measures the integrated squared difference between the forecast cumulative distribution function (CDF) $F(Q)$ and the observed discharge $Q^{\mathrm{obs}}$:
\begin{equation}
\mathrm{CRPS}(F, Q^{\mathrm{obs}}) = \int_{-\infty}^{\infty} [F(Q) - \mathbb{I}(Q \ge Q^{\mathrm{obs}})]^2 \, dQ,
\label{eq:crps}
\end{equation}
where $\mathbb{I}(Q \ge Q^{\mathrm{obs}})$ equals~1 if $Q \ge Q^{\mathrm{obs}}$ and~0 otherwise. 
For ensemble forecasts, this integral is approximated discretely as:
\[
\mathrm{CRPS}(F, Q^{\mathrm{obs}}) = \frac{1}{M} \sum_{m=1}^{M} |Q_m^{\mathrm{sim}} - Q^{\mathrm{obs}}| - \frac{1}{2M^2} \sum_{i=1}^{M}\sum_{j=1}^{M} |Q_i^{\mathrm{sim}} - Q_j^{\mathrm{sim}}|,
\]
where $M$ is the ensemble size. 
Lower CRPS values indicate forecasts that are both well-calibrated (statistically consistent with observations) and sharp (concentrated around the observed value), with the optimal score of~0 corresponding to a perfect forecast.

To assess the model's probabilistic behavior for high-flow events, we complement CRPS with reliability and sharpness diagnostics~\citep{murphy1973new,wilks2011statistical,jolliffe2012forecast}.
Reliability evaluates the agreement between forecast probabilities and observed event frequencies, while sharpness quantifies the concentration of forecast probabilities independent of observations. 
For high-flow events (top~10\% of streamflows), the ensemble forecast at each day~$t$ is converted to a probability of exceedance,
\[
p_t = P(Q_t^{\mathrm{sim}} > Q_{0.9}^{\mathrm{obs}}),
\]
where $Q_{0.9}^{\mathrm{obs}}$ is the 90th percentile of observed streamflow. 
Predicted probabilities $\{p_t\}$ are grouped into $B$ discrete bins (e.g., 0--0.1, 0.1--0.2, $\ldots$, 0.9--1.0), and for each bin~$b$, the mean predicted probability $\bar{p}_b$ and observed frequency $o_b$ yield the reliability score:
\begin{equation}
\mathrm{Reliability} = \frac{1}{B} \sum_{b=1}^{B} (\bar{p}_b - o_b)^2.
\label{eq:reliability}
\end{equation}
Perfect reliability occurs when $\bar{p}_b = o_b$ across all bins. 

Sharpness is computed as the variance of predicted probabilities,
\[
\mathrm{Sharpness} = \mathrm{Var}(p_t),
\label{eq:sharpness}
\]
where high sharpness indicates confident forecasts (probabilities near~0 or~1), and low sharpness indicates uncertain forecasts (probabilities near~0.5).

While reliability and sharpness evaluate distributional performance, PR analysis focuses on event detection skill. 
Precision is the fraction of predicted high-flow events that were correct, and recall measures the fraction of observed high-flow events successfully identified:
\begin{equation}
\mathrm{Precision} = \frac{TP}{TP + FP}, \qquad
\mathrm{Recall} = \frac{TP}{TP + FN},
\label{eq:precision_recall}
\end{equation}
where $TP$, $FP$, and $FN$ denote true positives, false positives, and false negatives. 
High precision indicates few false alarms, whereas high recall indicates strong event detection capability.
Finally, to quantify improvements relative to a reference model, we compute skill scores for NSE, KGE, and CRPS as:
\begin{align}
\mathrm{NSESS} &= \frac{\mathrm{NSE}_{\text{model}} - \mathrm{NSE}_{\text{ref}}}{1 - \mathrm{NSE}_{\text{ref}}}, \label{eq:nse_skill}\\
\mathrm{KGESS} &= \frac{\mathrm{KGE}_{\text{model}} - \mathrm{KGE}_{\text{ref}}}{1 - \mathrm{KGE}_{\text{ref}}}, \label{eq:kge_skill}\\
\mathrm{CRPSS} &= 1 - \frac{\mathrm{CRPS}_{\text{model}}}{\mathrm{CRPS}_{\text{ref}}}. \label{eq:crps_skill}
\end{align}
Positive skill scores indicate improvement of the target model over the reference, with an optimal value of~1 corresponding to perfect improvement and~0 indicating no change.

\subsubsection{Evaluation Strategy}

In the hydrologic validation phase, HydroDiffusion is compared against DiffusionLSTM\textsubscript{EncDec}, DiffusionLSTM\textsubscript{Dec}, DRUM, and the simulation benchmark using the metrics NSE, KGE, COR, PBias, FHV, and FLV. 
For each diffusion model (excluding DRUM), a deterministic version is also included to demonstrate the benefit of generative training; only their statistical performance is reported without further analysis. 
Because the released DRUM implementation supports only a single-step Day-0 simulation, comparisons are restricted to that setting. 
Temporal consistency for HydroDiffusion and DiffusionLSTM is further examined by evaluating predictive accuracy across zero- to seven-day lead times using NSE, KGE, and COR.

The streamflow reforecast phase then evaluates the relative performance of the SSM and LSTM backbones over one- to seven-day lead times using the skill scores NSESS, KGESS, and CRPSS. 
Spatial distributions of these skill scores are analyzed to identify regional variations in performance across CONUS. 
In addition, probabilistic diagnostics, including reliability--sharpness diagrams and precision--recall analyses, are used to assess ensemble calibration, forecast confidence, and flood event identification.

\section{Results}
\label{sec:results}

This section evaluates the performance of the proposed diffusion models across two complementary experimental phases. 
First, we examine the hydrologic validity of the models when driven by observed forcings. 
Second, we evaluate generalization under operational forecasting conditions using zero-shot reforecast experiments with GEFSv12 meteorological inputs. 
Together, these analyses establish both the physical consistency and practical forecasting skill of the proposed HydroDiffusion framework relative to LSTM-based baselines.

\begin{table}[!b]
\centering
\renewcommand{\arraystretch}{1.5} 
\caption{
Day-0 performance of all evaluated models across 531 CAMELS basins. Values for diffusion-based models are computed from the ensemble mean. For each metric, basin-level medians are shown with interquartile ranges (25th--75th percentiles). 
FHV (top 0.1\% flow) and FLV (bottom 30\% flow) are expressed as percentages, where values closer to zero indicate better performance. All other metrics are unitless, with higher values denoting better performance. The best median value in each column (excluding the simulation benchmark) is highlighted in red.
\label{tab:day0_main}}
\begin{tabularx}{\linewidth}{X c c c c c}
\toprule
Model & NSE~(↑) & KGE~(↑) & COR~(↑) & FHV~(\%~$\rightarrow$0) & FLV~(\%~$\rightarrow$0) \\
\midrule
 Simulation benchmark & $0.71^{+0.07}_{-0.11}$ & $0.74^{+0.08}_{-0.11}$ & $0.87^{+0.03}_{-0.05}$ & $-15.54^{+16.63}_{-13.58}$ & $-1.13^{+37.00}_{-76.76}$ \\
 \midrule
LSTM$_{\text{EncDec}}$ & $0.69^{+0.09}_{-0.14}$ & $0.74^{+0.07}_{-0.11}$ & $0.85^{+0.04}_{-0.06}$ & $-9.72^{+17.38}_{-17.43}$ & $\textcolor{red}{\boldsymbol{-1.41^{+40.13}_{-93.95}}}$\\
LSTM$_{\text{Dec}}$ & $0.68^{+0.09}_{-0.11}$ & $0.76^{+0.07}_{-0.12}$ & $0.85^{+0.05}_{-0.06}$ & $-8.80^{+18.53}_{-15.11}$ & $-15.81^{+49.99}_{-208.30}$ \\
SSM & $0.73^{+0.07}_{-0.11}$ & $0.76^{+0.07}_{-0.11}$ & $0.88^{+0.03}_{-0.06}$ & $-16.44^{+14.75}_{-13.23}$ & $7.76^{+32.16}_{-50.86}$ \\
 \midrule
DRUM$^{*}$ & $0.71^{+0.08}_{-0.11}$ & $0.76^{+0.08}_{-0.10}$& $0.87^{+0.04}_{-0.05}$ & $\textcolor{red}{\boldsymbol{-3.37^{+18.59}_{-17.89}}}$ & $19.77^{+25.68}_{-28.24}$ \\
DiffusionLSTM$_{\text{EncDec}}$ & $0.73^{+0.08}_{-0.11}$& $0.75^{+0.06}_{-0.11}$ & $0.87^{+0.04}_{-0.05}$ & $-17.18^{+14.74}_{-13.94}$ & $-2.02^{+37.97}_{-59.01}$\\
DiffusionLSTM$_{\text{Dec}}$ & $0.73^{+0.08}_{-0.12}$ & $0.73^{+0.08}_{-0.11}$ & $0.87^{+0.04}_{-0.06}$ & $-23.10^{+13.33}_{-13.27}$ & $-10.17^{+33.27}_{-84.04}$ \\
HydroDiffusion (ours) & $\textcolor{red}{\boldsymbol{0.75^{+0.07}_{-0.11}}}$& $\textcolor{red}{\boldsymbol{0.78^{+0.07}_{-0.09}}}$ & $\textcolor{red}{\boldsymbol{0.88^{+0.03}_{-0.06}}}$ & $-12.03^{+14.90}_{-13.76}$ & $9.88^{+33.08}_{-36.48}$ \\
\bottomrule
\end{tabularx}
\raggedright
{\footnotesize
$^{*}$Adopted from \cite{ou2025probabilistic}.}
\end{table}

\subsection{Model Hydrologic Validation with Observed Forcings}
\label{subsec:calibration_results}

\textbf{Day-0 Simulation Performance.}
To begin, we validate the diffusion models using their Day-0 performance against the state-of-the-art simulation benchmark across 531 watersheds (\Cref{tab:day0_main}), following the rationale discussed in Section~\ref{subsec:forecast_framework}. We also include deterministic counterparts of each diffusion model to isolate the benefits of generative training. Reported values are basin-level medians with interquartile ranges (25th--75th percentiles). All diffusion model performances are computed from the ensemble mean. 
The best-performing model for each metric (excluding the simulation benchmark) is highlighted in red.

All diffusion models achieve performance at least comparable to the simulation benchmark, with HydroDiffusion showing the highest accuracy, highlighting its improved ability to reproduce watershed hydrologic behavior.
HydroDiffusion outperforms its LSTM-based counterparts (DiffusionLSTM\textsubscript{EncDec} and DiffusionLSTM\textsubscript{Dec}) as well as DRUM in NSE, KGE, and COR, suggesting the advantage of the proposed diffusion framework and the SSM backbone. 
In contrast, DRUM achieves the lowest high-flow bias, indicating its strong capability in estimating extreme flow events. 
For further spatial basin-wise comparison between HydroDiffusion and DRUM, we include maps of skill scores in \Cref{fig:spatial_day0_drum} in Appendix. 
Between the two DiffusionLSTM variants, the encoder--decoder configuration consistently outperforms the decoder-only configuration, suggesting that LSTMs benefit from an explicit temporal encoding in diffusion-based modeling.

\begin{figure}[!t]
\centering
\includegraphics[width=\textwidth]{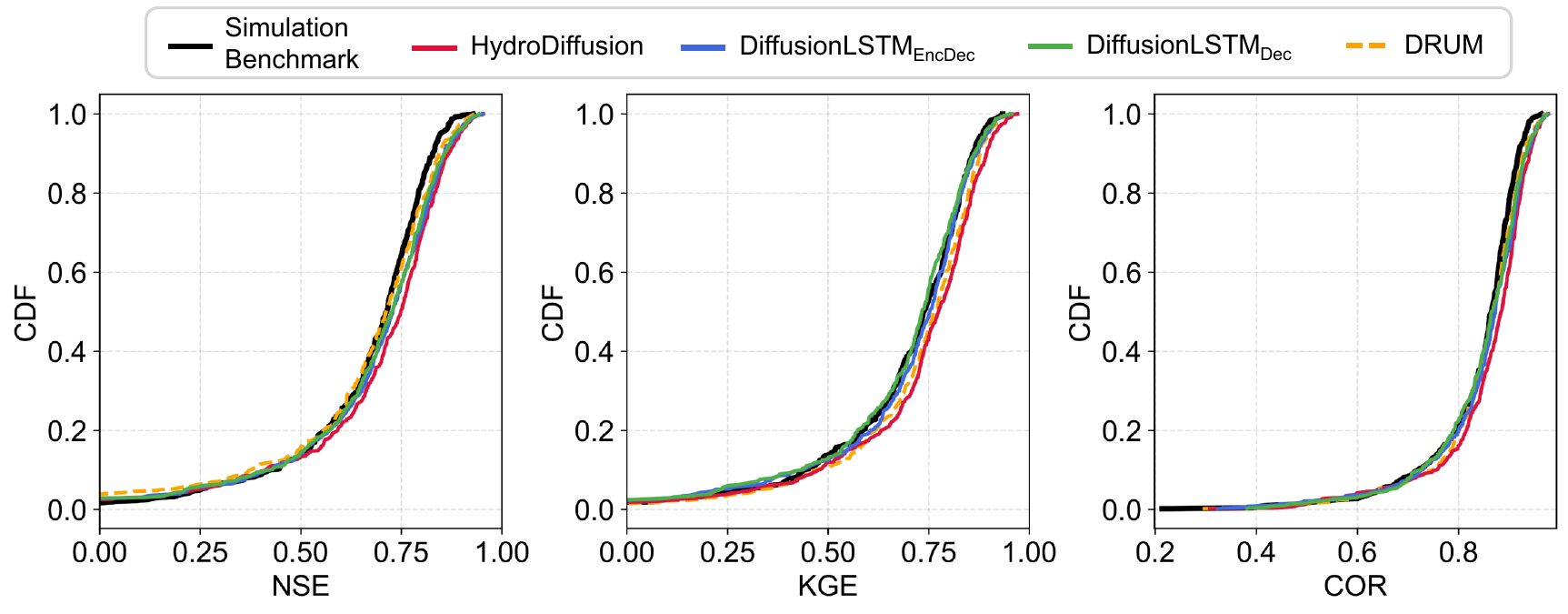}
\caption{
Cumulative distribution functions (CDFs) of basin-level deterministic performance metrics (NSE, KGE, and COR) across 531 CAMELS watersheds at Day-0. Results are shown for HydroDiffusion (red), DiffusionLSTM\textsubscript{EncDec} (blue), DiffusionLSTM\textsubscript{Dec} (green), and the DRUM baseline (yellow), with the deterministic simulation benchmark (black) serving as a reference.
}
\label{fig:cdf_plot}
\end{figure}

To complement the median and interquantile ranges, we present the cumulative distribution functions (CDFs) of NSE, KGE, and COR across 531 watersheds in \Cref{fig:cdf_plot}.
CDFs for diffusion models are computed from ensemble means. 
The simulation benchmark is shown in black, HydroDiffusion in red, DiffusionLSTM\textsubscript{EncDec} in blue, DiffusionLSTM\textsubscript{Dec} in green, and DRUM as a yellow dashed line. 
HydroDiffusion exhibits a rightward shift relative to the other diffusion models, confirming consistent outperformance across basins. 
DRUM yields the lowest NSE but slightly higher KGE than DiffusionLSTMs, while all three achieve similar COR values. 
Within the DiffusionLSTM group, DiffusionLSTM\textsubscript{EncDec} shows a higher KGE, reinforcing the advantage of the encoder--decoder structure for LSTM backbones.

\begin{figure}[!t]
\centering
\includegraphics[width=\textwidth]{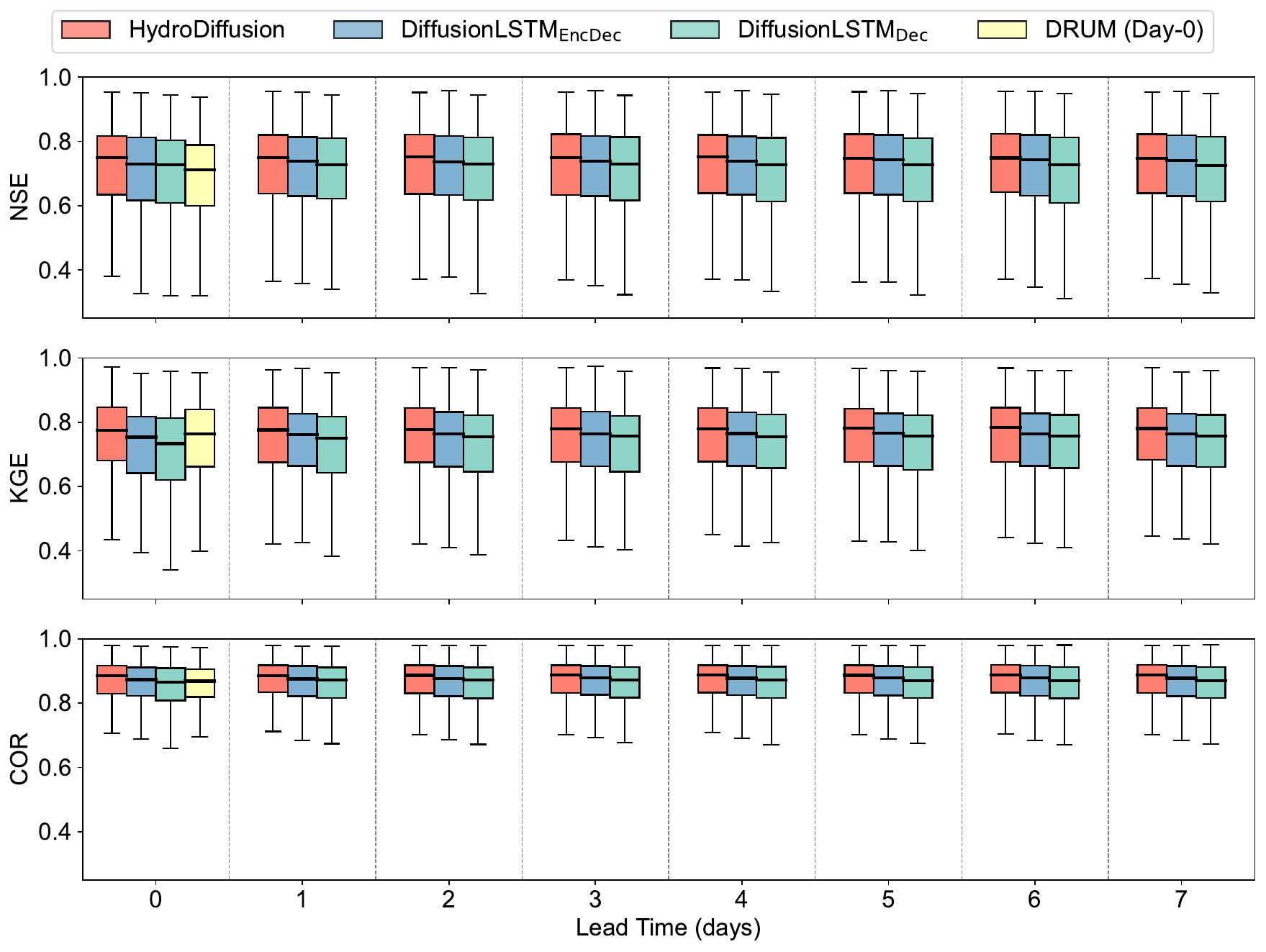}
\caption{
Boxplots of NSE, KGE, and COR across 531 CAMELS watersheds for HydroDiffusion (red), DiffusionLSTM\textsubscript{EncDec} (blue), DiffusionLSTM\textsubscript{Dec} (green), and the DRUM baseline (yellow) over the zero- to seven-day simulation horizon. 
The DRUM model is shown only for Day-0 performance.
}
\label{fig:boxplot_calibration_acrosslead}
\end{figure}

\textbf{Performance Validation across the Simulation Horizon.}
To assess whether the validated Day-0 performance persists throughout the simulation, we extend the comparison to a seven-day simulation horizon. \Cref{fig:boxplot_calibration_acrosslead} presents boxplots of NSE, KGE, and COR across 531 watersheds for the four diffusion models. 
Both HydroDiffusion and its LSTM-based counterparts maintain stable performance across the simulation horizon, indicating that model hydrologic validity holds without skill degradation. The relative performance among the backbones is also consistent. Specifically, HydroDiffusion outperforms the two DiffusionLSTMs, and the encoder--decoder version of DiffusionLSTM also continues to outperform the decoder-only version. 

\subsection{Comparison between SSM and LSTM Backbones}
\label{subsec:backbone_comparison}

To evaluate backbone generalization under operational forecasting conditions, we conduct a zero-shot reforecast experiment using GEFSv12 meteorological forcings.
Because the hydrologic validation phase showed that the encoder--decoder LSTM outperformed the decoder-only version, this comparison focuses on HydroDiffusion and DiffusionLSTM\textsubscript{EncDec}.

\begin{figure}[!t]
\centering
\includegraphics[width=\textwidth]{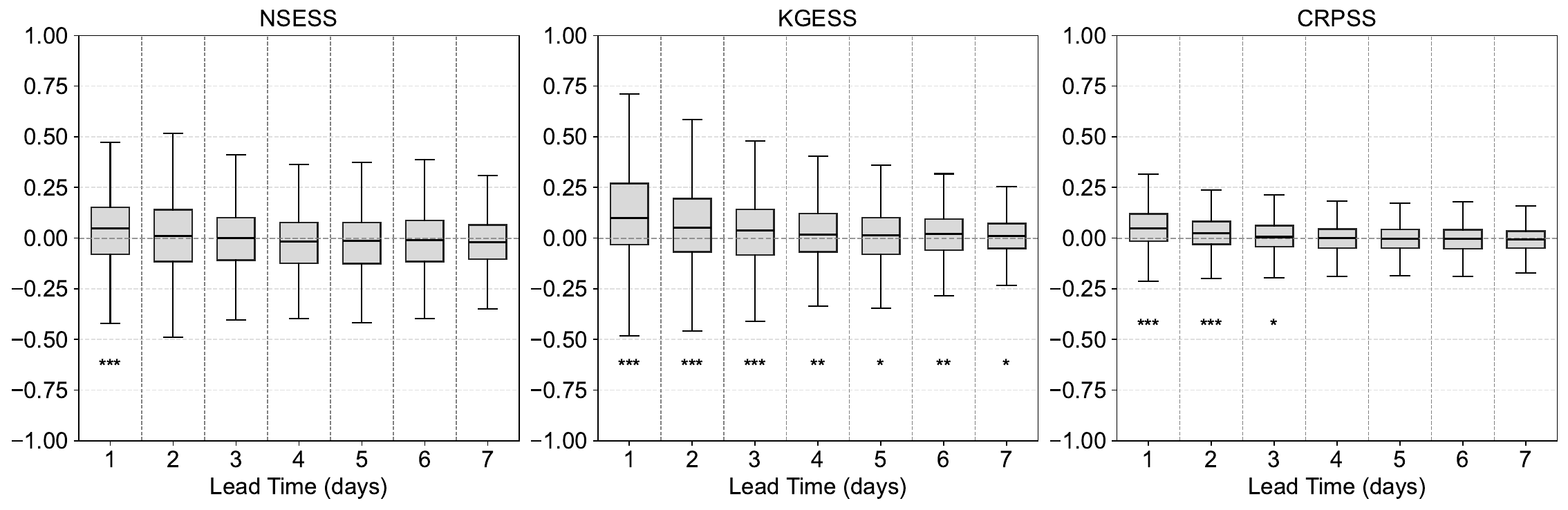}
\caption{Boxplots of NSESS, KGESS, and CRPSS for HydroDiffusion relative to DiffusionLSTM\textsubscript{EncDec} across 531 watersheds. Positive values indicate improved performance of HydroDiffusion compared to DiffusionLSTM\textsubscript{EncDec}. Statistical significance of the improvements is assessed using a one-sided Wilcoxon signed-rank test (\textit{H\textsubscript{1}}: skill~$>$~0), with significance levels denoted as *\,$p<0.05$, **\,$p<0.01$, and ***\,$p<0.001$.}
\label{fig:boxplot_zst_acrosslead}
\end{figure}

\textbf{Forecast Skill across Lead Times.}
We first present the relative performance of HydroDiffusion versus DiffusionLSTM\textsubscript{EncDec} across 531 watersheds across seven-day lead times, as shown in \Cref{fig:boxplot_zst_acrosslead}.
The Day-0 nowcast results are excluded, as they rely on observed forcings rather than true forecasts.
Positive values indicate better HydroDiffusion performance. 
Statistical significance is assessed using a one-sided Wilcoxon signed-rank test~\citep{wilcoxon1945individual} under the alternative hypothesis $H_1$:~skill~$>$~0, with significance levels denoted by * ($p<0.05$), ** ($p<0.01$), and *** ($p<0.001$).

Overall, the relative skill improvement of HydroDiffusion over the DiffusionLSTM\textsubscript{EncDec} gradually decreases with increasing lead time. 
Nevertheless, HydroDiffusion maintains a clear advantage in KGE, showing statistically significant improvements at all lead times. 
Using the NSE and CRPS metrics, the advantage remains significant up to one- and three-day leads, respectively. 
These results highlight that the SSM backbone yields higher forecast skill, particularly in the early lead times.

\begin{figure}[!t]
\centering
\includegraphics[width=\textwidth]{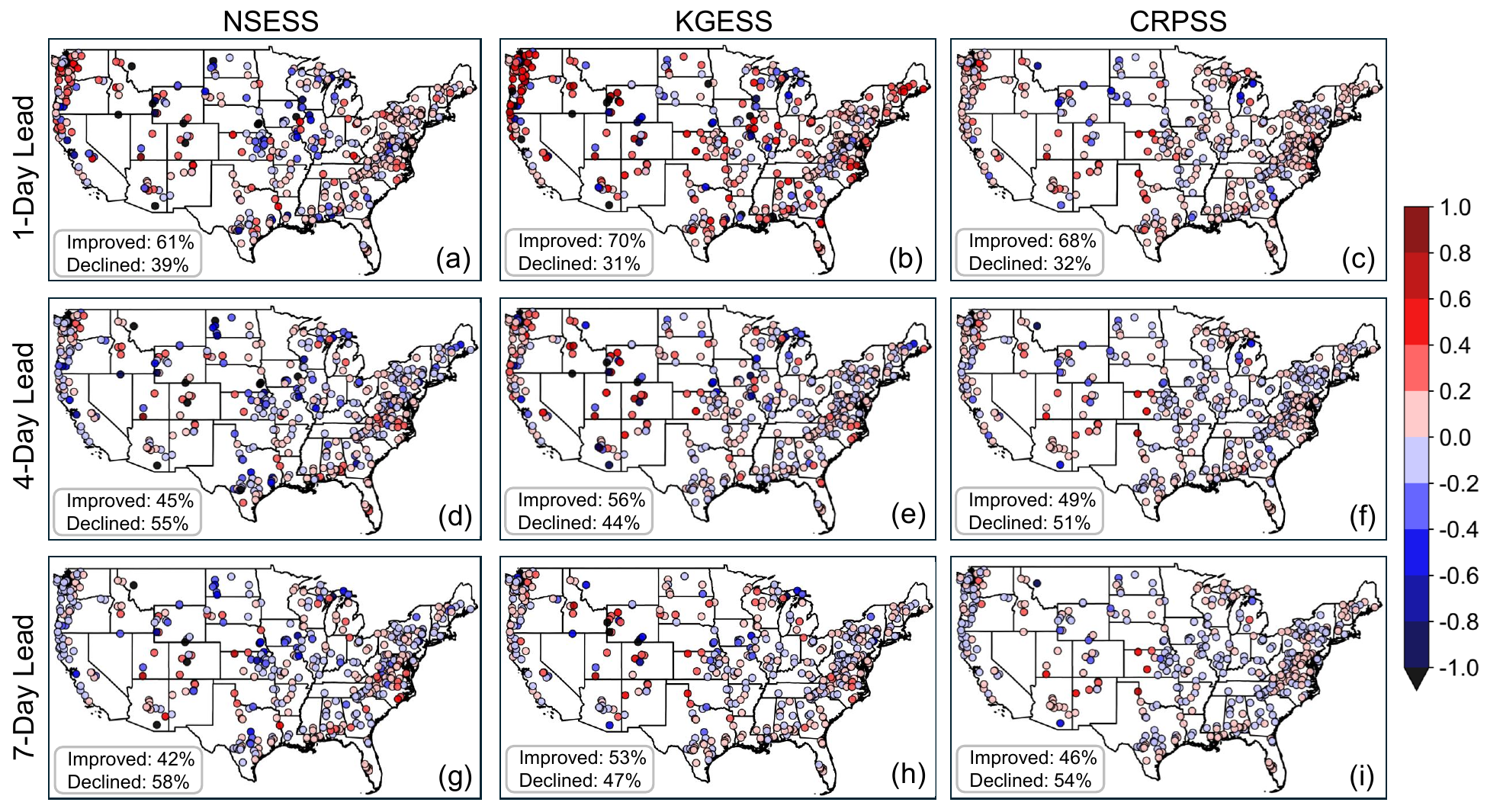}
\caption{
Spatial distributions of skill scores comparing HydroDiffusion and DiffusionLSTM\textsubscript{EncDec} across 531 watersheds for lead times of one, four, and seven days.
Each column corresponds to (a, d, g) NSESS, (b, e, h) KGESS, and (c, f, i) CRPSS.
Positive (red) values denote improved performance of HydroDiffusion relative to DiffusionLSTM\textsubscript{EncDec}, while negative (blue) values indicate reduced skill.
Percentages of watersheds with improved and declined performance are annotated in the lower-left corner of each panel. 
}
\label{fig:spatial_zst_acrosslead}
\end{figure}

\textbf{Forecast Skill Spatial Variability.}
\Cref{fig:spatial_zst_acrosslead} depicts the spatial distribution of NSESS, KGESS, and CRPSS for lead times of one, four, and seven days. 
At the one-day lead, HydroDiffusion shows widespread improved performance over the DiffusionLSTM\textsubscript{EncDec} in roughly 60--70\% of watersheds across all three metrics, demonstrating the advantage of the SSM backbone in forecasting.
Negative skill values occur mainly in the Appalachians, northern Great Plains, Great Lakes region, and isolated Rocky Mountain watersheds. 
KGESS exhibits the largest regional variability, with strong positive skill ($>0.6$) in the southern U.S. and Pacific Northwest but localized declines ($<-1$) in parts of Wyoming and Arizona.  In contrast, CRPSS varies more narrowly (approximately~$\pm$0.2), suggesting that the probabilistic metrics are spatially more consistent than the deterministic metrics.

For a four-day lead, the proportion of improved basins decreases to roughly 45--56\%. HydroDiffusion retains a modest advantage in KGE, while NSE and CRPS differences narrow. 
Negative skill values appear more frequently in the Midwest and Appalachian regions, whereas positive skill persists in the Pacific Northwest and the Rockies, indicating greater forecast stability in those areas. 
By a seven-day lead, the spatial pattern remains largely unchanged. 
Noticeable positive NSE skill ($>0.2$) is observed in parts of the eastern U.S. and scattered Rocky Mountain and Southwest basins. 
KGE improvements remain more extensive than NSE, particularly across Idaho, Wyoming, Colorado, and Utah. 
CRPSS again remains relatively stable (–0.2 to 0.2), indicating that the probabilistic performance of HydroDiffusion declines more gradually than deterministic accuracy with lead time. 

\begin{figure}[!t]
\centering
\includegraphics[width=\textwidth]{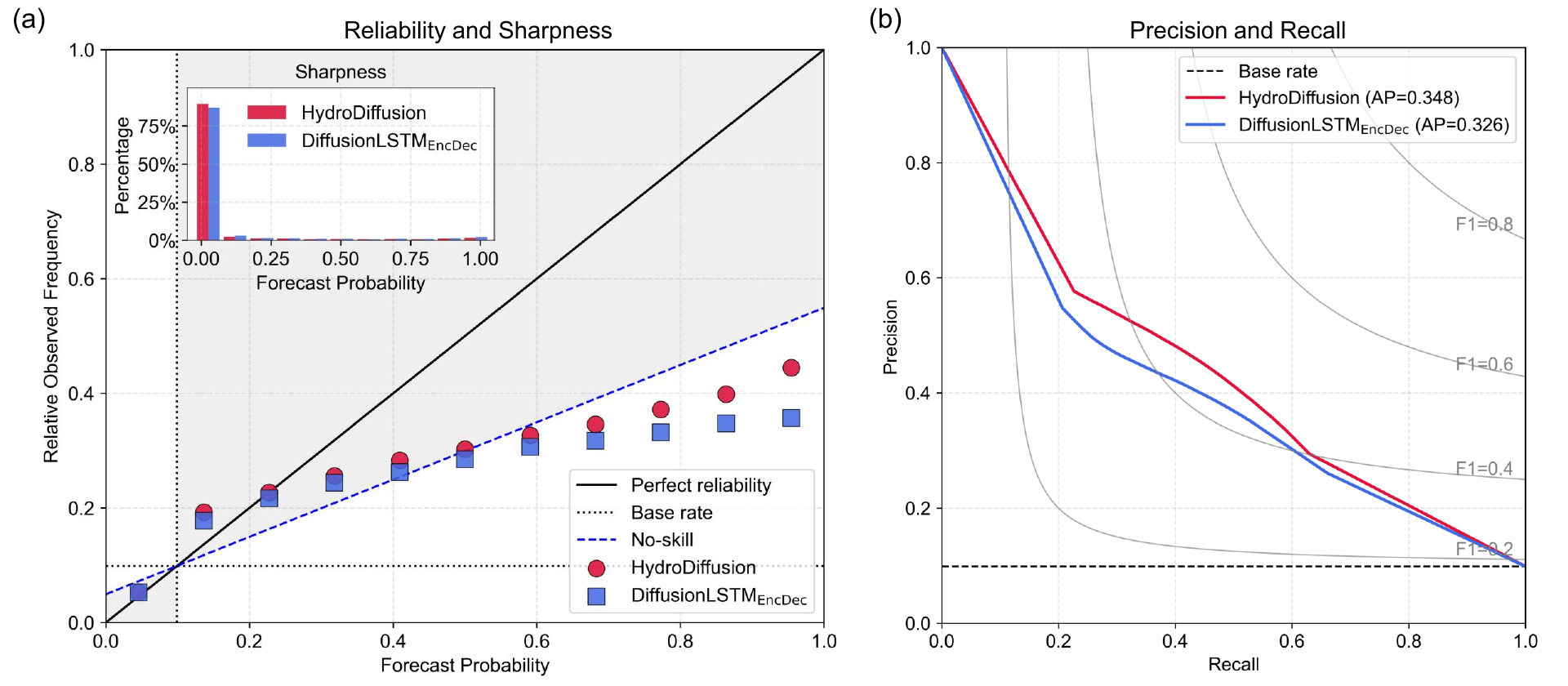}
\caption{
(a) Reliability diagram and corresponding sharpness distributions for HydroDiffusion (red) and DiffusionLSTM\textsubscript{EncDec} (blue), evaluated for the top 10\% high-flow regime aggregated across all lead times and 531 watersheds.
The solid black line denotes perfect reliability, the dashed blue line the no-skill reference, and the dotted black lines the climatological base rate.
The shaded region indicates the skillful zone relative to climatology, and the inset histogram shows forecast sharpness, expressed as the distribution of predicted probabilities.
(b) Precision–recall (PR) diagram comparing HydroDiffusion (red) and DiffusionLSTM\textsubscript{EncDec} (blue), also aggregated across all lead times and watersheds.
The dashed horizontal line marks the base event rate, while gray iso-contours represent constant F\textsubscript{1}-scores ($F_1 = 2PR/(P+R)$), with higher values indicating better trade-offs between precision and recall.
Average precision (AP) values are reported in parentheses.
}
\label{fig:reliability_diagram_zst}
\end{figure}

\textbf{Assessment of Forecast Skill for High-Flow Events.}
To further examine ensemble behavior, \Cref{fig:reliability_diagram_zst} presents reliability--sharpness (panel~a) and precision--recall (panel~b) analyses for the top-10\% high-flow regime, aggregated across all lead times and basins. 
In \Cref{fig:reliability_diagram_zst}(a), both models show general overconfidence—the predicted exceedance probabilities exceed observed frequencies—evidenced by points below the 1:1 line. Skill relative to climatology is maintained up to the probability bin of 0.6, after which reliability degrades and falls outside the skillful zone. Across all bins, HydroDiffusion remains closer to the diagonal than DiffusionLSTM\textsubscript{EncDec}, indicating more calibrated probabilistic forecasts. 
The inset sharpness histogram reveals that most forecast probabilities cluster near zero for both models, which is as expected for rare high-flow events. However, both models exhibit modest sharpness, implying a tendency to produce conservative probability estimates rather than confident forecasts for extreme flows.

\Cref{fig:reliability_diagram_zst}(b) further evaluates high-flow detection skill using precision--recall curves. 
HydroDiffusion achieves consistently higher precision across a broad range of recall values, with its curve (red) lying above that of DiffusionLSTM\textsubscript{EncDec} (blue). 
Consequently, HydroDiffusion attains a higher average precision (AP) score of 0.348 compared to 0.326 for DiffusionLSTM\textsubscript{EncDec}, confirming its superior discrimination of high-flow events.

\section{Discussion}
\label{sec:discussion}

This section discusses the effectiveness of the joint denoising strategy and SSM backbone within the diffusion framework for probabilistic streamflow forecasting, followed by key challenges, limitations, and directions for future research.

\subsection{Model Design, Validation, and Forecast Evaluation}

A major contribution of this study is the development of a diffusion-based generative model specifically tailored for streamflow forecasting. 
While diffusion models were originally introduced for high-dimensional image generation~\citep{ho2020denoising,song2020score}, and later extended to two-dimensional hydrometeorological applications such as precipitation forecasting~\citep{gao2023prediff,wang2024skillful,nai2024reliable,zhong2024fuxi,sha2025improving}, their recursive frame-by-frame design in those domains is motivated by memory constraints rather than by sequence dynamics. 
In contrast, basin-scale hydrologic forecasting operates in a lower-dimensional space, making it computationally feasible to jointly denoise the entire multi-day trajectory in a single pass. 
Building on this property, we design HydroDiffusion to generate full streamflow trajectories simultaneously, enhancing temporal coherence across lead times and mitigating the stepwise error accumulation typical of recursive denoising approaches~\citep{ning2023input,ma2024utsd}.

The hydrologic validation results (\Cref{tab:day0_main}, \Cref{fig:cdf_plot}–\Cref{fig:boxplot_calibration_acrosslead}) confirm the effectiveness of this design. 
HydroDiffusion achieves higher accuracy than DRUM~\citep{ou2025probabilistic} for Day-0 simulation and maintains consistent skill across the seven-day simulation horizon, indicating physically plausible hydrologic behavior and temporal stability. 
Beyond establishing model validity, these results provide a transparent, CONUS-wide benchmark under observed forcings for future probabilistic forecasting research.

Under the operational reforecast setup with NOAA GEFSv12 forcings, HydroDiffusion demonstrates robust generalization without retraining. 
Compared to its strongest LSTM-based counterpart (i.e., DiffusionLSTM\textsubscript{EncDec}), HydroDiffusion achieves statistically significant gains in KGE, CRPS, and NSE up to lead times of seven, three, and one days, respectively (\Cref{subsec:backbone_comparison}). 
Spatially, HydroDiffusion outperforms across most regions, with underperformance concentrated in the Appalachians, northern Great Plains, Great Lakes, and parts of the Rockies. 
Notably, strong skill (NSE, KGE, CRPS~$>0.4$) persists in the Rockies—including Wyoming, Utah, and Colorado—even under the more uncertain reforecast conditions. 
Probabilistic diagnostics (\Cref{fig:reliability_diagram_zst}) further show that HydroDiffusion produces more reliable and better-calibrated ensembles, indicating improved discrimination of high-flow events and overall robustness to forcing uncertainty.

Despite these advances, the high-flow forecasting skill for both HydroDiffusion and DiffusionLSTM\textsubscript{EncDec} remains limited (\cref{fig:reliability_diagram_zst}). 
Both models tend to be overconfident relative to climatology, underestimating event frequency and failing to capture the full observed variability. 
This behavior likely results from underprediction of extreme flows and underdispersive ensemble spreads~\citep{legg2004early,zhao2022two}. 
Improving ensemble calibration and rare-event representation, therefore, remains essential for operational deployment.

\subsection{Challenges, Limitations, and Future Work}
While HydroDiffusion generally outperforms DiffusionLSTM\textsubscript{EncDec}, its regional variability raises an important question: do the observed gains truly reflect architectural advantages, or could they partly result from random alignment with errors in the hydrometeorological forecasts?  
Because streamflow forecasting operates downstream of atmospheric processes, uncertainty arises not only from the hydrologic model but also from inaccuracies in meteorological inputs, and, critically, from their interactions. 
This makes it difficult to attribute performance differences solely to model structure~\citep{troin2021generating,zhang2025alternative}.  

Evidence of this coupling appears in our results: HydroDiffusion exhibits declining reforecast skill in the Appalachian and Pacific Northwest regions from one- to seven-day leads (\Cref{fig:spatial_zst_acrosslead}), whereas its validation performance under observed forcings remains stable across the simulation horizon (\Cref{fig:boxplot_calibration_acrosslead} and~\Cref{fig:spatial_hydro_val_across_horizon} in Appendix).  
The discrepancy likely reflects uncertainty propagation from the GEFSv12 reforecast forcings and the nonlinear interactions between model dynamics and input errors.
One contributor to this propagated uncertainty could be the mismatch in forcing resolution. The coarse GEFSv12 reforecasts differ significantly from the high-resolution Daymet forcings used during training.
When these reforecasts are bilinearly interpolated to basin scales, regional biases can arise, particularly in mountainous regions where complex terrain and orographic processes generate fine-scale meteorological variability~\citep{schreiner2020impact}.
Future research should explicitly disentangle uncertainty sources—meteorological, hydrologic, and structural—using established frameworks for joint uncertainty analysis~\citep{ajami2007integrated,vrugt2008treatment,gong2013estimating,montanari2008estimating}.  
Such analyses would clarify the spatial patterns of skill and guide targeted model improvements.

A second limitation lies in the absence of physically based forecast benchmarks.  
Although deep learning models often surpass traditional hydrologic models in predictive accuracy~\citep{nearing2024global,hunt2022using}, they lack comparable diagnostic interpretability.  
Physically based models allow event-level reasoning, linking forecast errors to hydrologic processes such as infiltration or snowmelt~\citep{clark2017evolution,paniconi2015physically}, whereas deep networks remain largely opaque.  Developing physically based reforecast archives would enable systematic cross-paradigm comparisons, offering both reference baselines and interpretive context for evaluating deep learning forecasts. 

Looking ahead, integrating physical constraints within diffusion models represents a promising path toward greater interpretability and realism.  
However, such hybridization should be purpose-driven and region-specific rather than applied uniformly.  
Rigid constraints, such as the precipitation-based mass-conservation condition in the MC-LSTM~\citep{hoedt2021mc}, can degrade performance in certain hydrologic regimes, whereas adaptive formulations improve accuracy~\citep{wang2025investigating,wang2025mass}.  
Targeted strategies, such as emphasizing snow dynamics in snow-dominated catchments~\citep{wang2025towards} or incorporating event-based constraints in precipitation-driven basins~\citep{wang2025deep} may yield interpretable gains without sacrificing flexibility.

Finally, advancing deep probabilistic hydrologic forecasting will require coordinated community standards and evaluation protocols, similar to those established for deterministic simulation~\citep{kratzert2019toward,kratzert2019towards,frame2022deep,liu2024probing,wang2025deep}.  
Shared benchmarks with standardized forcings, lead times, metrics (e.g., CRPS, reliability, sharpness, discrimination), and watershed selections would promote transparency, reproducibility, and foster progress toward reliable and physically interpretable hydrologic forecasting systems.

\section{Conclusions}
\label{sec:conclusions}
This study introduces HydroDiffusion, a diffusion-based probabilistic streamflow forecasting framework that combines modern score-based diffusion with a decoder-only SSM backbone. 
Compared to prior work, we propose three key modifications for diffusion-based probabilistic hydrologic forecasting. 
First, we adopt a continuous-time score-based formulation for training. 
Second, we advocate jointly denoising the entire forecast trajectory rather than predicting only the next element and unrolling the model autoregressively. 
Third, we replace recurrent LSTMs with a high-expressivity SSM backbone. 
Together, these advances enable HydroDiffusion to produce temporally coherent and physically consistent ensemble forecasts while maintaining high predictive skill across lead times.

Beyond methodological innovation, this work shows the importance of establishing strong baselines in hydrologic machine learning. 
Our experiments reveal that even a deterministic SSM backbone outperforms previously proposed diffusion frameworks, highlighting the critical role of model architecture in both deterministic and probabilistic forecasting skill. 
These findings suggest that meaningful progress in probabilistic hydrologic modeling depends not only on adopting generative approaches but also on pairing them with competitive state-of-the-art network architectures for sequence modeling.

Overall, we establish a new modeling paradigm for probabilistic hydrologic forecasting. 
It demonstrates how the generative diffusion framework can be combined with advanced sequence modeling architectures to capture both hydrologic uncertainty and temporal structure in a unified formulation. 
By integrating modern diffusion learning with efficient state space modeling, this work provides a foundation for next-generation probabilistic forecasting systems in hydrology and Earth system science.


\section{Open Research}
NOAA Global Ensemble Forecast System version 12 (GEFSv12) Re-forecast is available at \url{https://registry.opendata.aws/noaa-gefs-reforecast}. The Catchment Attributes and Meteorology for Large-sample Studies (CAMELS) dataset is available in \cite{newman_2022_15529996}.
Python codes to replicate the results of this study are available at \url{https://github.com/yhwang08/HydroDiffusion}. 

\section*{Acknowledgment}
This work is supported by the U.S. Department of Energy, Office of Science, Office of Advanced Scientific Computing Research, Scientific Discovery through Advanced Computing (SciDAC) program, under Contract Number DE-AC02-05CH11231 at Lawrence Berkeley National Laboratory. 
Partial support was provided by the Lab Directed Research and Development (LDRD) program at Berkeley Lab. 
The U.S. Government retains, and the publisher, by accepting the article for publication, acknowledges, that the U.S. Government retains a non‐exclusive, paidup, irrevocable, world‐wide license to publish or reproduce the published form of this manuscript, or allow others to do so, for U.S. Government purposes. 
%

\bibliographystyle{plainnat}
\bibliography{Reference}


\clearpage
\appendix
\section{Appendix}

\begin{table}[!h]
\centering
\caption{Hydrometeorological variables and static catchment attributes for model training.}
\label{tab:variables}
\resizebox{0.85\textwidth}{!}{%
\begin{tabular}{clp{9cm}}
\toprule
\textbf{ID} & \textbf{Variable} & \textbf{Description} \\
\midrule
\multicolumn{3}{l}{\textit{Hydrometeorological Variables}} \\
1 & Maximum air temp & 2 m daily maximum air temperature (°C) \\
2 & Minimum air temp & 2 m daily minimum air temperature (°C) \\
3 & Precipitation & Average daily precipitation (mm/day) \\
4 & Radiation & Surface-incident solar radiation (W/m\textsuperscript{2}) \\
5 & Vapor pressure & Near-surface daily average vapor pressure (Pa) \\
\midrule
\multicolumn{3}{l}{\textit{Static Catchment Attributes}} \\
6  & Precipitation mean & Mean daily precipitation \\
7  & PET mean & Mean daily potential evapotranspiration \\
8  & Aridity index & Ratio of mean PET to mean precipitation \\
9  & Precip seasonality & Annual seasonality of precipitation estimated by fitting sine curves; positive (negative) values indicate summer (winter) peak; near zero indicates uniform distribution \\
10 & Snow fraction & Fraction of precipitation falling on days with temperature $<$ 0°C \\
11 & High precipitation frequency & Frequency of days with precipitation $\geq$ 5× mean daily precipitation \\
12 & High precip duration & Mean duration of high precipitation events (consecutive days with $\geq$ 5× mean daily precipitation) \\
13 & Low precip frequency & Frequency of dry days ($<$ 1 mm/day) \\
14 & Low precip duration & Mean duration of dry periods (consecutive days with precipitation $<$ 1 mm/day) \\
15 & Elevation & Catchment mean elevation \\
16 & Slope & Catchment mean slope \\
17 & Area & Catchment area \\
18 & Forest fraction & Fraction of catchment covered by forest \\
19 & LAI max & Maximum monthly mean of leaf area index \\
20 & LAI difference & Difference between the maximum and minimum mean of the leaf area index \\
21 & GVF max & Maximum monthly mean of green vegetation fraction \\
22 & GVF difference & Difference between the maximum and minimum monthly mean of green vegetation fraction \\
23 & Soil depth (Pelletier) & Depth to bedrock (maximum 50 m) \\
24 & Soil depth (STATSGO) & Soil depth (maximum 1.5 m) \\
25 & Soil porosity & Volumetric porosity \\
26 & Soil conductivity & Saturated hydraulic conductivity \\
27 & Max water content & Maximum water content of the soil \\
28 & Sand fraction & Fraction of sand in the soil \\
29 & Silt fraction & Fraction of silt in the soil \\
30 & Clay fraction & Fraction of clay in the soil \\
31 & Carbonate rocks fraction & Fraction of catchment area characterized as “carbonate sedimentary rocks” \\
32 & Geological permeability & Surface permeability (log$_{10}$ scale) \\
\bottomrule
\end{tabular}
}
\end{table}
\clearpage

\begin{table}[!h]
\centering
\caption{Training hyperparameters and configuration for DiffusionLSTM (for both encoder--decoder and decoder-only configurations) and its deterministic counterpart.}
\label{tab:lstm_hparams}
\resizebox{0.85\textwidth}{!}{%
\begin{tabular}{c p{3.2cm} c c}
\toprule
\textbf{ID} & \textbf{Parameter} & \textbf{DiffusionLSTM} & \textbf{LSTM} \\
\midrule
1  & \texttt{hidden\_size}               & 256 & 256 \\
2  & \texttt{batch\_size}                & 256 & 256 \\
3  & \texttt{initial\_forget\_gate\_bias} & 3   & 3   \\
4  & \texttt{learning\_rate}             & $3\times10^{-5}$ (warmup+decay) &
\begin{tabular}[t]{@{}l@{}}
$1\times10^{-3},\;5\times10^{-4},\;1\times10^{-4}$ \\
(10, 10, 10 epochs)
\end{tabular} \\
5  & \texttt{dropout}                    & 0.5 & 0.4 \\
6  & \texttt{epochs}                     & 60  & 30  \\
7  & \texttt{optimizer}                  & Lion & Adam \\
8  & \texttt{loss function}               & MSE  & NSE \\
\bottomrule
\end{tabular}
}
\end{table}

\begin{table}[!h]
\centering
\caption{Training hyperparameters and configuration for HydroDiffusion (S4D-FT).}
\label{tab:hydrodiffusion_hparams}
\resizebox{0.85\textwidth}{!}{%
\begin{tabular}{cllp{9cm}}
\toprule
\textbf{ID} & \textbf{Parameter} & \textbf{Employed Value} & \textbf{Description} \\
\midrule
1  & \texttt{d\_model}       & 256          & Number of channels per S4D layer. \\
2  & \texttt{d\_state}       & 256          & Number of latent SSM states per channel. \\
3  & \texttt{n\_layers}      & 6            & Number of stacked S4D layers. \\
4  & \texttt{cfr}            & 10.0         & Real-part scale of the state matrix $A$ (S4D-FT only). \\
5  & \texttt{cfi}            & 10.0         & Imaginary-part scale of the state matrix $A$ (S4D-FT only). \\
6  & \texttt{ssm\_dropout}   & 0.2          & Dropout applied within S4D layers.\\
7  & \texttt{min\_dt}        & 0.01         & Minimum time step in discretization. \\
8  & \texttt{max\_dt}        & 0.1          & Maximum time step in discretization. \\
9  & \texttt{epochs}         & 60           & Number of training epochs. \\
10 & \texttt{batch\_size}    & 256          & Training samples per batch. \\
11 & \texttt{learning\_rate} & $3\times10^{-5}$ & Peak global learning rate (after warm-up); applied to non-SSM parameters with linear warm-up and linear decay. \\
12 & \texttt{lr\_min}        & $3\times10^{-6}$ & Learning rate used for S4D kernel parameters $(A,B,C,\Delta t)$; the effective S4D LR is $\min(\texttt{learning\_rate}, \texttt{lr\_min})$. \\
13 & \texttt{lr\_dt}         & 0.001        & Learning rate applied exclusively to the time-step parameter $\Delta t$ within each S4D kernel. \\
14 & \texttt{weight\_decay}  & 0.0          & Global $L_2$ regularization for non-SSM parameters. \\
15 & \texttt{wd}             & $4\times10^{-5}$ & $L_2$ regularization applied to S4D kernel parameters. \\
16 & \texttt{optimizer}      & Lion         & Optimizer used during training. \\
17 & \texttt{loss\_target}   & MSE          & Training objective for score learning. \\
\bottomrule
\end{tabular}
}
\end{table}
\clearpage

\begin{table}[!h]
\centering
\caption{Training hyperparameters and configuration for SSM (i.e., S4D-FT).}
\label{tab:ssm_hparams}
\resizebox{0.85\textwidth}{!}{%
\begin{tabular}{cllp{9cm}}
\toprule
\textbf{ID} & \textbf{Parameter} & \textbf{Employed Value} & \textbf{Description} \\
\midrule
1  & \texttt{d\_model}       & 128          & Number of channels per S4D layer. \\
2  & \texttt{d\_state}       & 128          & Number of latent SSM states per channel. \\
3  & \texttt{n\_layers}      & 6            & Number of stacked S4D layers. \\
4  & \texttt{cfr}            & 10.0         & Real-part scale of the state matrix $A$ (S4D-FT only). \\
5  & \texttt{cfi}            & 10.0         & Imaginary-part scale of the state matrix $A$ (S4D-FT only). \\
6  & \texttt{ssm\_dropout}   & 0.12          & Dropout applied within S4D layers.\\
7  & \texttt{min\_dt}        & 0.01         & Minimum time step in discretization. \\
8  & \texttt{max\_dt}        & 0.1          & Maximum time step in discretization. \\
9  & \texttt{epochs}         & 50           & Number of training epochs. \\
10 & \texttt{batch\_size}    & 256          & Training samples per batch. \\
11 & \texttt{learning\_rate} & $4\times10^{-4}$ & Peak global learning rate (after warm-up); applied to non-SSM parameters with linear warm-up and linear decay. \\
12 & \texttt{lr\_min}        & $4\times10^{-5}$ & Learning rate used for S4D kernel parameters $(A,B,C,\Delta t)$; the effective S4D LR is $\min(\texttt{learning\_rate}, \texttt{lr\_min})$. \\
13 & \texttt{lr\_dt}         & 0.001        & Learning rate applied exclusively to the time-step parameter $\Delta t$ within each S4D kernel. \\
14 & \texttt{weight\_decay}  & 0.03          & Global $L_2$ regularization for non-SSM parameters. \\
15 & \texttt{wd}             & 0.02 & $L_2$ regularization applied to S4D kernel parameters. \\
16 & \texttt{optimizer}      & Lion         & Optimizer used during training. \\
17 & \texttt{loss function}   & NSE          & Training objective for score learning. \\
\bottomrule
\end{tabular}
}
\end{table}
\clearpage

 \begin{figure}[!h]
 \centering
 \includegraphics[width=\textwidth]{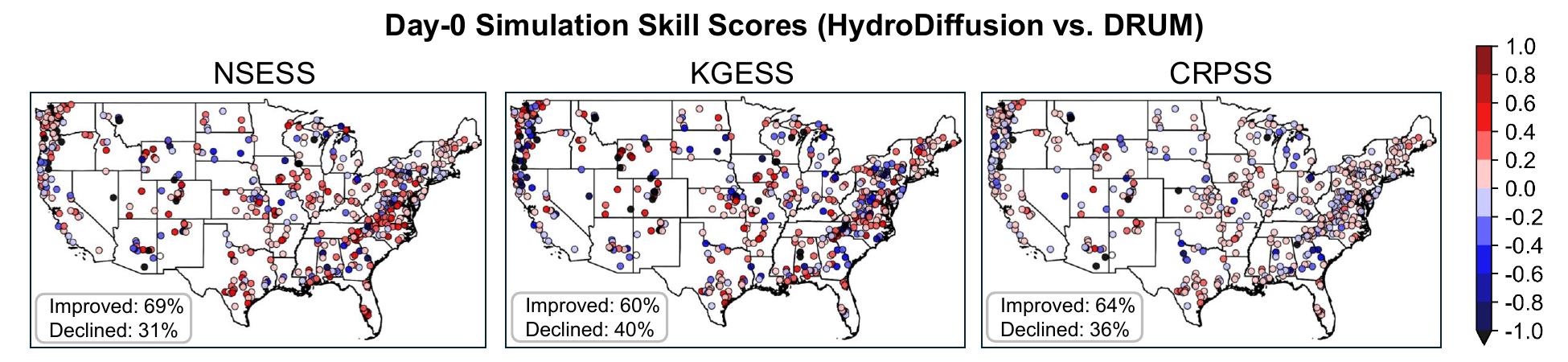}
 \caption{
 Spatial distributions of skill scores comparing HydroDiffusion and DRUM across 531 watersheds during the hydrologic validation phase using Daymet forcings at Day-0.
 Positive (red) values denote improved performance of HydroDiffusion relative to DRUM, while negative (blue) values indicate reduced skill.
 }
 \label{fig:spatial_day0_drum}
 \end{figure}

 \begin{figure}[!h]
 \centering
 \includegraphics[width=\textwidth]{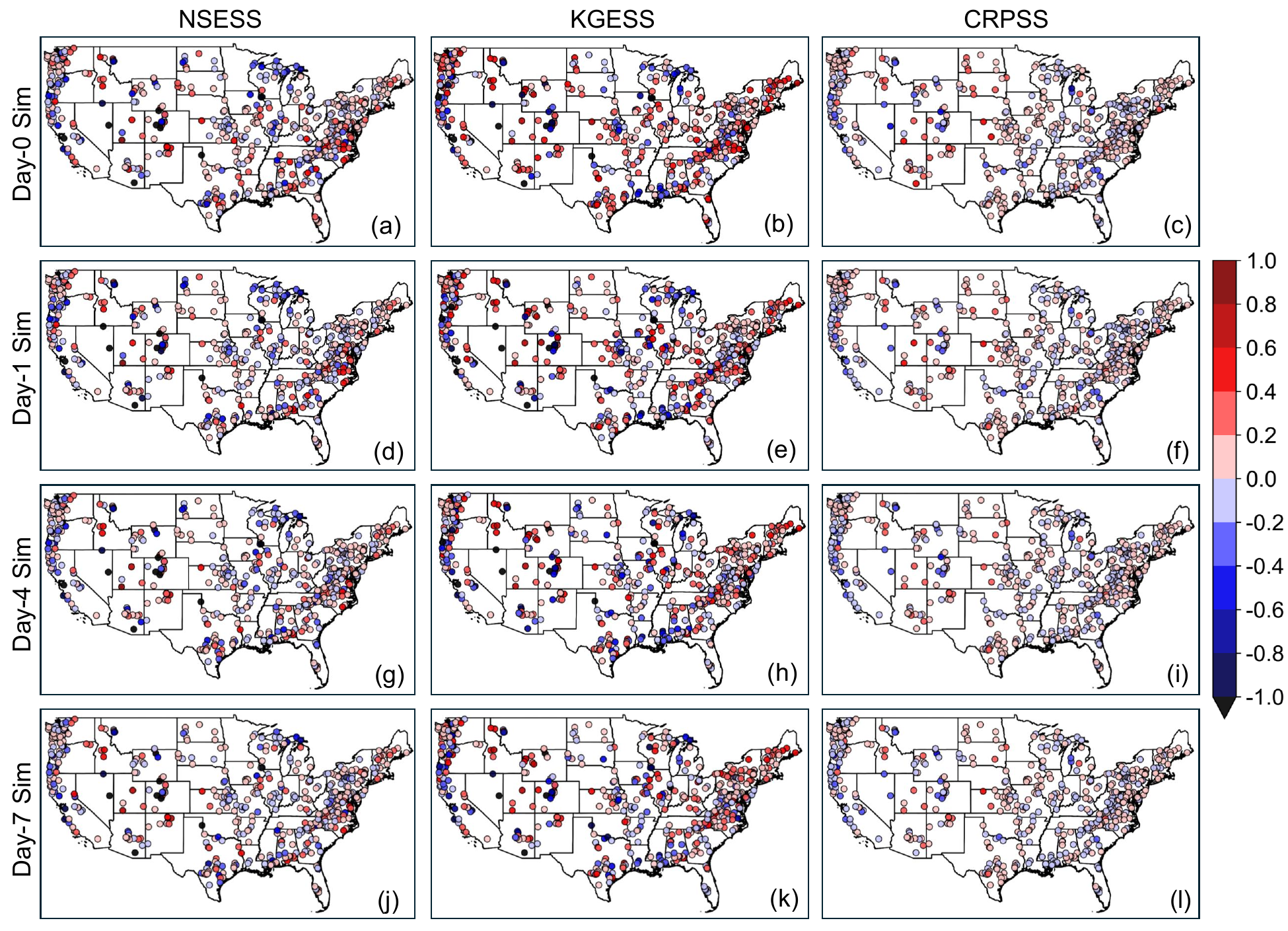}
 \caption{
 Spatial distributions of skill scores comparing HydroDiffusion and DiffusionLSTM\textsubscript{EncDec} across 531 watersheds during the hydrologic validation phase using Daymet forcings.
  Results are shown for simulation days zero, one, four, and seven. 
 Each column corresponds to (a, d, g) NSESS, (b, e, h) KGESS, and (c, f, i) CRPSS.
 Positive (red) values denote improved performance of HydroDiffusion relative to DiffusionLSTM\textsubscript{EncDec}, while negative (blue) values indicate reduced skill.
 }
 \label{fig:spatial_hydro_val_across_horizon}
 \end{figure}

\end{document}